\documentclass[11pt]{article}

\usepackage[final]{acl}

\usepackage{times}
\usepackage{latexsym}
\usepackage[most]{tcolorbox}
\usepackage{amsmath}
\usepackage[inline]{enumitem}
\usepackage[T1]{fontenc}

\usepackage[utf8]{inputenc}
\usepackage[table]{xcolor}
\usepackage{microtype}

\usepackage{inconsolata}
\usepackage{booktabs}
\usepackage{graphicx}
\usepackage{todonotes}
\usepackage{cleveref}
\usepackage{fontenc}[T1]
\usepackage{tipa}

\usepackage{booktabs}
\usepackage{multirow}
\usepackage{graphicx}
\usepackage{adjustbox}
\usepackage{longtable}

\usepackage{tcolorbox}
\tcbuselibrary{listings, breakable, skins}
\usepackage{xcolor}

\definecolor{queryblue}{RGB}{30,90,160}
\definecolor{toolgreen}{RGB}{30,130,80}
\definecolor{concludered}{RGB}{160,50,30}
\definecolor{outputgold}{RGB}{160,115,0}
\definecolor{lightblue}{RGB}{235,243,255}
\definecolor{lightgreen}{RGB}{235,250,240}
\definecolor{lightred}{RGB}{255,240,237}
\definecolor{lightyellow}{RGB}{255,252,225}
\definecolor{framegray}{RGB}{80,80,100}

\tcbset{
steptcb/.style={
enhanced, sharp corners,
boxrule=0.5pt, arc=2pt,
left=5pt, right=5pt, top=2pt, bottom=2pt,
fontupper=\footnotesize,
}
}

\newcommand{\cb}[2][]{\Note[#1]{CB}{cyan!20}{#2}}

\newcommand{\Note}[4][]{\todo[author=#2,color=#3,fancyline,size=\small,#1]{#4}}

\newcommand{\jz}[1]{\textcolor{red}{\bf\small [#1 --Jian]}}
\newcommand{\tfc}{\textsc{Typological Feature Coding}}
\newcommand{\tht}{\textsc{Typological Hypothesis Testing}}
\newcommand{\agent}{\textsc{AutoTypologist}}

\title{LLM Agents as Computational Typologists}

\author{Changbing Yang\textsuperscript{\textipa{\OE}}, Christopher Hammerly\textsuperscript{\textipa{\OE}}, Freda Shi\textsuperscript{\textipa{W},\textipa{V}}, Jian Zhu\textsuperscript{\textipa{\OE}} \\
\textsuperscript{\textipa{\OE}}University of British Columbia\\
\textsuperscript{\textipa{W}}University of Waterloo \quad 
\textsuperscript{\textipa{V}}Vector Institute \\
  \texttt{cyang33@mail.ubc.ca},  \texttt{jian.zhu@ubc.ca}
  }

\usepackage{pifont}

\newcommand{\cmark}{\textcolor{green!60!black}{\ding{51}}}
\newcommand{\xmark}{\textcolor{red!70!black}{\ding{55}}}

\tcbuselibrary{skins, breakable, listings}
\usepackage{xcolor}
 
\usepackage{array}
\usepackage{booktabs}
\usepackage{tikz}
\usetikzlibrary{arrows.meta}
 
\definecolor{queryblue}{RGB}{30,90,160}
\definecolor{toolgreen}{RGB}{30,130,80}
\definecolor{concludered}{RGB}{160,50,30}
\definecolor{outputgold}{RGB}{180,130,0}
\definecolor{lightblue}{RGB}{235,243,255}
\definecolor{lightgreen}{RGB}{235,250,240}
\definecolor{lightred}{RGB}{255,240,237}
\definecolor{lightyellow}{RGB}{255,252,230}
\definecolor{auditgray}{RGB}{245,245,248}
\definecolor{steelblue}{RGB}{70,130,180}
\usepackage{tabularx}

\begin{document}

\maketitle
\begin{abstract}
Linguistic typology relies on expert analysis of reference grammars across languages, making large-scale crosslinguistic comparison labor-intensive and unscalable. We introduce \agent{}, an LLM agent for evidence-grounded typological analysis over reference grammars. The agent is capable of retrieving relevant grammar sections, analyzing interlinear glossed text (IGT), and iteratively reasoning over typological hypotheses using a ReAct-style workflow. We evaluate the system on \tfc{} against expert annotations and \tht{} with typological universals using 25 open-source reference grammars. Operating under different information constraints in \tfc{}, the agent can synthesize information from reference grammar prose but still faces challenges with only IGTs in the target language. In \tht{}, the agent can synthesize crosslinguistic evidence and identify both supporting cases and counterexamples. These findings suggest that LLM agents can support scalable and inspectable typological analysis, while still requiring expert validation.

\end{abstract}

\section{Introduction}
Linguistic typology seeks to uncover linguistic universals of human language \cite{greenberg1963some,greenberg2005language,comrie1989language}. Typological knowledge is typically derived from comparing, re-analyzing, and synthesizing crosslinguistic data collected through linguistic fieldwork. 
Typological research leads to open-sourced typological linguistic databases, such as the World Atlas of Language Structures (WALS) \cite{wals}, the Atlas of Pidgin and Creole Language Structures (APiCS) \cite{michaelis2013atlas}, Grambank \cite{skirgardGrambankRevealsImportance2023}. In these databases, grammar is decomposed into a few hundred individual features, where each language will be assigned a value for its grammatical structure. Traditionally,  creating such a database takes years of effort from hundreds of trained linguists, which is not scalable. 

\begin{figure*}[t]
    \centering
    \includegraphics[width=0.8\textwidth]{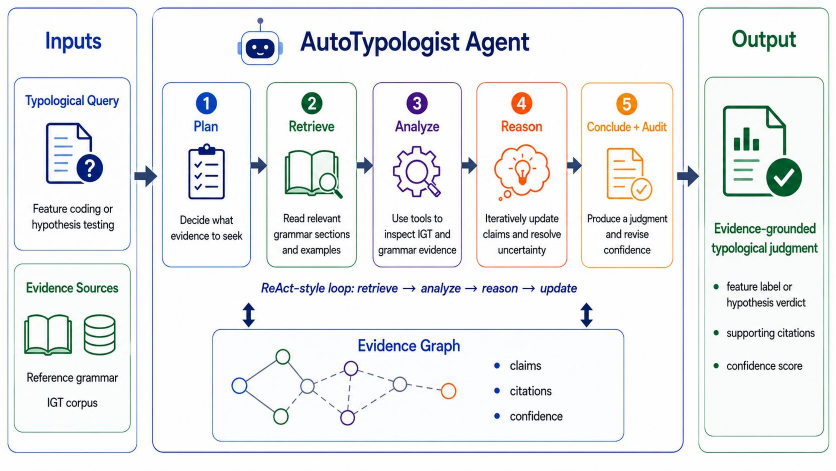}
    \caption{Simplified AutoTypologist pipeline. The agent converts a typological query into an evidence-grounded judgment by planning what evidence is needed, retrieving relevant grammar sections and IGT examples, analyzing the retrieved evidence with linguistically motivated tools, and iteratively updating its claims through a ReAct-style loop. A structured evidence graph records claims, citations, and confidence scores, supporting transparent feature coding or hypothesis testing.}
    \label{fig:autotypologist-pipeline}
\end{figure*}

Recent research suggests that LLMs are capable of performing meta-linguistic analyzes in languages \cite{sahin-etal-2020-puzzling,bean2024lingoly,begus2025large,yang-etal-2025-linggym,schneider2026speaking}, making them potential general problem solvers in language documentation and linguistic typology.

To accelerate and support expert typological analysis, as well as motivated by these pioneering works, we present \agent{}, an LLM agent 
that can actively query reference grammars to summarize and compare the grammatical structures of typologically diverse languages. 
Our contributions are: First, we introduce the \agent{}, the first LLM agent for evidence-grounded typological analysis. Our research opens up new opportunities to large-scale typological analyses, as LLM agents can assist linguists in gathering evidence for hypothesis testing.

Second, we formalize two agentic typology tasks: \tfc{}, where the agent assigns typological feature values and is evaluated against expert annotations, and \tht{}, where it tests typological universals by synthesizing crosslinguistic evidence. \agent{} reproduces expert codings with high agreement when reference-grammar prose is available and identifies both supporting evidence and counterexamples in hypothesis testing.

Third, we release a typological dataset built from 25 open-source reference grammars across 14 language families, containing structured grammar chunks, extracted interlinear glossed text (IGT) examples, and metadata for evaluating evidence-grounded typological reasoning.

Our research departs from the established NLP approach of developing task-specific and language-specific models but instead pursues abstract language-agnostic meta-linguistic reasoning, where LLMs \textit{``communicate explicit conceptualizations of language phenomena"} \cite{schneider2026speaking}. We believe \agent{} points to a promising direction for human-AI collaboration in linguistic research.



\section{Background}

\subsection{LLM agents for auto-research}
In recent years, researchers are actively exploring using LLM agents to accelerate scientific discoveries across disciplines, including mathematics, physics, and biochemical sciences \cite{romera2024mathematical,merchant2023scaling,fawzi2022discovering,hayes2025simulating,zhang2024scientific}. Generally, LLMs exhibit potential to perform scientific research, or at least a part of the workflow \cite{schmidgall-etal-2025-agent}, but still struggle to learn scientific taste \cite{tong2026ai}.
In contrast, relatively few studies have explored how LLM agents can conduct research in the social sciences and humanities \cite{grossmann2023ai,bail2024can,ziems2024can}, but the interest is rising rapidly, especially in sociology, management, and economics \cite{hao2025multi,ji2025automating,hu2026soclitgen,engzell2026paper,korinek2025ai}. However, despite the potential, LLM agents have yet to be applied to linguistic research, where low-resource and unseen languages can pose significant challenges. Our research aims to fill in the gap by developing a general LLM agent that can perform automated linguistic analysis.

\subsection{LLMs for meta-linguistic reasoning}
Some early works explore the automatic induction of grammatical rules from glossed texts \cite{zamaraeva-2016-inferring,howell-etal-2017-inferring,chaudhary-etal-2020-automatic,chaudhary-etal-2021-wall}. These approaches were built on traditional machine learning approaches and often center on specific aspects of grammar. 

Recently, studies find that incorporating metalinguistic knowledge from reference grammars can aid machine translation in low-resource languages \cite{tanzer2024benchmark,hus-anastasopoulos-2024-back}, and LLMs do possess abilities to understand meta-linguistic knowledge \cite{sahin-etal-2020-puzzling,bean2024lingoly,begus2025large,yang-etal-2025-linggym,schneider2026speaking}, though their understanding might be limited \cite{aycock2025can,marmonier-etal-2025-explicit}. Yet it has been observed that, at the moment, no fully agentic research workflows has been incorporated into language documentation or typological research workflow \cite{moeller2026computational,schneider2026speaking}. Rather than focusing on specific domains or languages, we develop a general LLM research agent that can perform autonomous typological research with meta-linguistic knowledge across languages and grammatical domains.

\section{Tasks}
We focus on two tasks essential to linguistic typology, with a formulation mirroring the actual research process. 

\paragraph{\tfc{}}
This task mirrors the feature coding process in Grambank \cite{skirgardGrambankRevealsImportance2023}, the largest typological database covering 2,467 languages in 195 features. 

The 195 features were curated by experienced linguists and include feature definitions, coding procedures, and illustrative evidence from source grammars. Feature GB020\footnote{\url{https://grambank.clld.org/parameters/GB020\#2/21.0/151.7}} below illustrates this format: feature values are assigned based on grammatical descriptions and examples from the relevant source materials (Examples that may be presented in interlinear glossed text (IGT) format. We explain the IGT format in \S\ref{igt-explain}). While the task instructions were originally developed for human linguists, they can also be adapted to LLM agents. 
Given the feature query and coding procedures, the agent is instructed to selectively read through a reference grammar to perform feature coding. The agent then produces both a predicted label and an evidence-grounded justification.

\begin{tcolorbox}[
enhanced,
colback=red!4,
colframe=red!35!black,
boxrule=0.5pt,
arc=2mm,
left=3mm,
right=3mm,
top=1.5mm,
bottom=1.5mm,
]
\textbf{Feature GB020.}
\textit{Are there definite or specific articles?}

\vspace{0.1em}

\vspace{0.1em}
\textbf{Procedure.}
\vspace{-0.8em}
\begin{enumerate}\setlength\itemsep{0pt}\setlength\parskip{0pt}
\item Code \textbf{1} if there is a morpheme that can mark definiteness or specificity without also conveying a spatial deictic meaning. 
\item Code \textbf{0} if no definite article is mentioned and none can be found in examples or texts in an otherwise comprehensive grammar. 
\item Code \textbf{?} if the grammar lacks enough analysis to determine whether a definite article exists.
\end{enumerate}
\textbf{Output.}\\
\begin{tabular}{l}
Language: Aiton \\
Coding label: \textbf{1} \\
Source: \newcite{morey2004tai}
\end{tabular}
\medskip
\begin{tabular}{lllll}
m\textschwa & nan& a m\={a}t& n\textepsilon & w\={a} \\
time& that & minister & DEF & say\\
\multicolumn{5}{l}{\textit{`Then the minister said.'}} \\
\end{tabular}
\end{tcolorbox}

\paragraph{\tht{}}
A central goal of linguistic typology is to formulate and evaluate hypotheses about crosslinguistic generalizations. Traditionally, human linguists test such hypotheses by comparing descriptions of many languages, identifying recurring structural patterns, and checking whether potential counterexamples exist. For example, the following classic typological hypothesis of word order is from \citet{greenberg1963some}.

\begin{tcolorbox}[
enhanced,
colback=red!4,
colframe=red!35!black,
boxrule=0.5pt,
arc=2mm,
left=3mm,
right=3mm,
top=1.5mm,
bottom=1.5mm,
]
\textbf{Universal 1. Word Order:}
\textit{In declarative sentences with nominal subject and object, the dominant order is almost always one in which the subject precedes the object.}
\end{tcolorbox}

Unlike \tfc{}, which focuses on assigning a feature
value to an individual language, \tht{} synthesizes evidence across multiple languages to evaluate cross-linguistic generalizations.
Hypothesis testing requires the agent to determine whether each language provides supporting evidence, counter-evidence, irrelevant evidence, or insufficient evidence, and then synthesize these language-level judgments into a crosslinguistic conclusion. 



\section{Typological Data}

\subsection{Reference grammars}
Our grammar corpus\footnote{Our data is available: \url{https://github.com/changbingY/typology-autosearch}} was constructed from \textit{Grammar Library},\footnote{\url{https://www.aaronbraver.com/resources/grammars/}} an online list of available reference grammars. We only selected open-sourced reference grammars with a permissive license and included in Grambank, mostly from published books and dissertations. In total, we included 25 reference grammars, representing 14 language families (see details in Appendix \ref{sec:all-reference-grammar-info}). 

\subsection{Preprocessing}
\label{igt-explain}
We digitized the collected reference grammars. First, each grammar was converted from PDF or \LaTeX{} format into plain text. For PDF-based grammars, we primarily used \texttt{pdfplumber}. For grammars available in \LaTeX{}, we directly extracted the textual content from the source files. 

We also extracted interlinear glossed text (IGT) samples, linguistic examples in which a target language sentence is accompanied by morpheme-by-morpheme segmentation, grammatical glosses (tags), and a free translation. The extracted materials were subsequently organized
into structured records with document-level and
example-level metadata for downstream retrieval and agent interaction, as described in Section~4.3. Below is an IGT example from Kagayanen \cite{pebley2024grammar}.\footnote{\textsc{T.R-clean-APL} = transitive realis ‘clean’ with applicative; \textsc{1S.ABS} = first-person singular absolutive, \textsc{3S.ERG} = third-person singular ergative, and \textsc{NABS} = non-absolutive.}
\begin{center}
\begin{adjustbox}{width=\columnwidth}
\begin{tabular}{ll}
\textbf{Source:} & Palimpyuan a din ta aparador. \\
\textbf{Morpheme:} & Pa-limpyo-an a din ta aparador. \\
\textbf{Gloss:} & T.R-clean-APL 1S.ABS 3S.ERG NABS cabinet \\
\textbf{Translation:} & `S/he cleaned the cabinet for me.' \\
\end{tabular}
\end{adjustbox}
\end{center}

IGT provides rich linguistic annotations for the agent to reason over grammatical categories of empirical language data. Candidate IGT examples were identified automatically by detecting numbered example labels. To ensure data quality, all digitized reference grammars and extracted IGT examples were manually reviewed by trained linguists to correct extraction errors.

\subsection{Indexing} To address context-length limitation, the digitized reference grammars were converted into structured chunks with unique identifiers. Each grammar was organized by chapter, sections, subsections, and subsubsections based on the original book. To make the representation suitable for retrieval, chunks were kept relatively small, typically capped at 1,000 words with meta information. Each chunk also included a short 2-3 sentence summary generated with \texttt{GPT5.4-mini} \cite{openai2026gpt54mini}, allowing the model to first access a brief overview before reading the full text of relevant sections. 
Each extracted IGT example was also stored as a structured entry with meta information.
Example chunks and IGT are shown in Appendix \ref{Grammar Chunk and IGT entry example}.

\section{AutoTypologist Framework}

\subsection{Agent workflow}

Our framework is a multi-stage LLM agent architecture for typological research.\footnote{Our code is available: \url{https://github.com/changbingY/typology-autosearch}} The system combines structured grammar reading, retrieval, quantitative IGT analysis,\footnote{To help the model interpret gloss tags, we also provide the gloss abbreviation list from each grammar as input.} evidence tracking, and iterative reasoning. Representative agent prompts are provided in Appendix~\ref{app:alltemp}. The agentic workflow was built using the ReAct framework \citep{yao2023react}. As illustrated in Fig.~\ref{fig:autotypologist-pipeline}, the workflow consists of the following stages:

\textbf{Research Planning.}
Queries of typological phenomena can be supplied explicitly by the user.\footnote{The system also supports automatic summarization based on the grammar structure; we describe this function in Appendix~\ref{app:auto-summary}.} The agent first reviews the reference grammar overviews\footnote{The grammar overview consists of the table of contents reconstructed from chapter, section, subsection, and subsubsection titles, together with summaries associated with each structured chunk.} and aggregate IGT statistics, and then constructs a per-query investigation plan for execution. 

\textbf{Targeted Pre-retrieval.}
Before the iterative reasoning loop begins, the agent executes the planned retrieval in two passes. 
First, it retrieves the identified grammar
sections and reads them in full, rather than relying on isolated text
chunks, and extracts explicit author claims into the evidence graph.
Second, it runs a set of planned IGT operations, including tag frequency
analysis, construction search, and absence checks over the IGT corpus.

\textbf{Iterative ReAct Investigation.}
The agent then performs a ReAct-style investigation loop \citep{yao2023react} in which
additional retrieval actions can be issued based on the evidence
accumulated so far. In our experiments, each language-level analysis was set up to 10 interaction rounds. At each round, the agent observes the current evidence state, reasons about what information is still missing or uncertain, and selects a tool call (\S\ref{tools}) such as retrieving additional grammar chunks, or terminates the loop by issuing a \textit{conclude} action. 

\textbf{Evidence Graph.}
Throughout the pre-retrieval and ReAct phases, the collected observations are incrementally organized into a structured evidence graph, where each node represents a one-sentence claim with meta information including evidence type, source, confidence score, and support/refute decision. The graph tracks supporting evidence, counter-evidence, uncertainty, and contradiction relations, while preserving links to the corresponding grammar chunks and IGT examples. 

\textbf{Synthesis and Reporting.}
Once the ReAct loop terminates, the agent produces an evidence-grounded typological analysis. The output includes the predicted feature value or typological conclusion, a justification grounded in the retrieved grammar chunks and IGT examples. Finally, the model re-examines the synthesized conclusion against the collected evidence. The auditor (using the same LLM with a separate prompt) can uphold, weaken, or overturn the conclusion, and assigns a revised confidence score. \footnote{We further examine the contribution of this auditing
stage through an ablation study on Aguaruna in Appendix~\ref{app:audit}.}

\subsection{Tools}
\label{tools}
\begin{table*}[t]
\centering
\small
\begin{tabular}{p{0.24\linewidth} p{0.18\linewidth} p{0.50\linewidth}}
\toprule
\textbf{Tool} & \textbf{Type} & \textbf{Function} \\
\midrule
\rowcolor{blue!5} \texttt{read\_full\_section} & Grammar & Read a complete grammar section. \\
\rowcolor{blue!5}\texttt{follow\_cross\_references} & Grammar & Follow internal references to related sections. \\
\rowcolor{blue!5}\texttt{extract\_author\_claims} & Grammar & Extract explicit analytical claims from prose. \\
\rowcolor{blue!5}\texttt{search\_text} & Grammar & Retrieve relevant grammar passages by hybrid search. \\

\rowcolor{green!5}\texttt{analyse\_tag} & IGT & Profile frequency, position, and co-occurrence of a tag. \\
\rowcolor{green!5}\texttt{analyse\_construction} & IGT & Find ordered gloss-tag sequences. \\
\rowcolor{green!5}\texttt{analyse\_absence} & IGT & Check whether a category is unattested or rare. \\
\rowcolor{green!5}\texttt{compare\_tags} & IGT & Compare two tags distributionally. \\
\rowcolor{green!5}\texttt{get\_section\_igt} & IGT & Retrieve IGT examples from matching sections. \\
\rowcolor{green!5}\texttt{search\_translations} & IGT & Search translation lines for semantic evidence. \\
\rowcolor{green!5}\texttt{get\_triline\_examples} & IGT & Retrieve aligned form, gloss, and translation examples. \\

\rowcolor{gray!10}\texttt{get\_tag\_inventory} & IGT-only & List all gloss tags by frequency. \\
\rowcolor{gray!10}\texttt{get\_construction\_inventory} & IGT-only & List frequent gloss n-gram patterns. \\
\rowcolor{gray!10}\texttt{find\_tag\_cluster} & IGT-only & Find tags that co-occur with a seed tag. \\
\rowcolor{gray!10}\texttt{analyse\_semantic\_context} & IGT-only & Inspect translation contexts for a tag. \\
\rowcolor{gray!10}\texttt{analyse\_morpheme\_position} & IGT-only & Infer prefix/suffix/stem position of a tag. \\
\rowcolor{gray!10}\texttt{get\_morpheme\_forms} & IGT-only & List surface forms associated with a tag. \\
\rowcolor{gray!10}\texttt{analyse\_tag\_usage} & IGT-only & Infer a tag’s function from examples. \\
\rowcolor{gray!10}\texttt{parse\_example\_structure} & IGT-only & Analyze clause structure in examples. \\
\bottomrule
\end{tabular}
\caption{Tools available to the agent during ReAct-style typological reasoning.}
\label{tab:agent-tools}
\end{table*}

We design a set of linguistically motivated tools to constrain the agent action space, each corresponding to a specific evidential operation a human typologist might perform in the research workflow. As summarized in Table~\ref{tab:agent-tools}, the tools fall into three broad groups:

\textbf{- Grammar-reading tools}
Grammar-reading tools operate over structured grammar chunks. 
These tools are intended to approximate the way a typologist consults a reference grammar: first locating the relevant descriptive section, then checking whether the author makes an explicit analytical claim about the target phenomenon.

\textbf{- IGT-analysis tools}
IGT-analysis tools operate over the extracted interlinear glossed examples to provide both aggregate and example-level evidence. 
These tools allow the agent to test whether grammatical categories described in grammar prose are also reflected in the observed examples, and to infer patterns directly from IGT data when prose descriptions are incomplete.

\textbf{- IGT-only exploratory tools}
In the IGT-only condition, the agent must reason with only IGT examples. We therefore provide additional exploratory tools designed to support grammatical induction from annotated data, while making the evidence used for each inference explicit and inspectable.

\textbf{Tool availability} is determined by the information sources accessible to the agent: grammar-reading tools are only available when reference grammar text is provided, IGT-only exploratory tools are only available when no reference grammar is present, and IGT-analysis tools are available whenever IGT examples are provided.

\textbf{Observability and literature grounding}
To enhance agent observability, we keep all agent traces during the multistep inferences. In each inference step, we force the agent to ground their decision by citing the unique identifier of original grammar chunks and IGTs from the corpus. This practice allows end users to inspect the agent decisions with reference to the sources to understand how agent decisions evolve.

\section{Experiments}

\paragraph{Models} We deployed our agents with a range of LLMs, including open-source models: \texttt{Qwen2.5-7B} \cite{qwen2.5}, \texttt{Qwen3.5-9B} \cite{qwen3.5}, \texttt{Qwen3.6-27B} \cite{qwen3.6-27b}, and \texttt{Gemma4-31B} \cite{gemma4_model_card}, and a closed-source model: \texttt{GPT5.4-mini} \cite{openai2026gpt54mini}. 
All open-sourced local models were deployed with \texttt{vllm} for efficient inference \cite{kwon2023efficient} in \texttt{bfloat16} precision. Full details are available in Appendix~\ref{app: hyper}.


\paragraph{Design}
In both \tfc{} and \tht{}, we contrast three different information conditions in which the agent has access to different sources of linguistic evidence:

\begin{center}
\begin{adjustbox}{width=\columnwidth}
\begin{tabular}{lcc}
\hline
\textbf{Condition} & \textbf{Grammar chunks} & \textbf{IGT entries} \\
\hline
\textbf{IGT} & \xmark & \cmark \\
\textbf{Grammar} & \cmark & \xmark \\
\textbf{Grammar+IGT} & \cmark & \cmark \\
\hline
\end{tabular}
\end{adjustbox}
\end{center}

The \textbf{IGT} condition presents a more challenging but realistic scenario in which only annotated linguistic examples are available, requiring the agent to infer grammatical features from raw data without access to author explanations. In the \textbf{Grammar} condition, the agent reasons over structured grammar chunks. The \textbf{Grammar+IGT} condition provides the richest setting, allowing the agent to combine explicit grammatical descriptions with example-based evidence from IGT entries.

For \tfc, we use all 25 languages coded in Grambank. For \tht, we select 10 languages per hypothesis-testing run: Aguaruna, Ch\'{a}cobo, Chakali, Moloko, Ik, Eastern Geshiza, Papuan Malay, Komnzo, Kalamang, and Pite Saami. We choose these languages to maximize coverage across language families. We selected 10 relatively general universals from Greenberg's original set of 45 universals, prioritizing hypotheses that could be evaluated across a broad range of reference grammars. We also manually created adversarial hypotheses by permuting or negating the original hypotheses, making them unlikely to be correct or to have appeared verbatim in pretraining data. The full list of selected universals and their adversarial counterparts is provided in Appendix~\ref{greenberg-all-rules-adv}. We evaluated three models for this task: \texttt{Qwen3.6-27B}, \texttt{Gemma4-31B}, and \texttt{GPT5.4-mini}. For each information condition, we ran the agent 3 times to account for variability. We report results averaged across these three runs. 
\vspace{-0.5em}
\paragraph{Baselines} 

Specifically for \tfc, we include two baselines for comparison. The first is a \textbf{zero-shot prompting} baseline, in which the model directly predicts feature codings without any additional contextual or grammatical guidance. This baseline evaluates whether LLMs can infer typological features of low-resource languages from pretrained knowledge alone. The second is a \textbf{majority-vote} baseline derived from Grambank, where each feature is assigned the value that occurs most frequently across languages (Details are shown in Appendix \ref{majority-voting-details}). Although simple, this baseline is difficult to beat: because Grambank feature-value distributions are highly imbalanced, majority voting benefits from strong dataset-level typological biases and can achieve competitive performance by favoring frequent values. Thus, improvements over the majority-vote baseline provide evidence that the model is capturing grammatical information beyond the majority-class prior.

\begin{table*}[t]
\centering
\small
\setlength{\tabcolsep}{4pt}
\begin{adjustbox}{width=\textwidth}
\begin{tabular}{l|cc|cc|cc|cc|cc|cc}
\toprule
\rowcolor{gray!20}
& \multicolumn{2}{c|}{\textbf{Majority Vote}}
& \multicolumn{2}{c|}{\texttt{Qwen2.5-7B}}
& \multicolumn{2}{c|}{\texttt{Qwen3.5-9B}}
& \multicolumn{2}{c|}{\texttt{Qwen3.6-27B}}
& \multicolumn{2}{c|}{\texttt{Gemma4-31B}}
& \multicolumn{2}{c}{\texttt{GPT5.4-mini}} \\
\rowcolor{gray!20}
\textbf{Condition}
& \textbf{Wgt.} & \textbf{Mac.}
& \textbf{Wgt.} & \textbf{Mac.}
& \textbf{Wgt.} & \textbf{Mac.}
& \textbf{Wgt.} & \textbf{Mac.}
& \textbf{Wgt.} & \textbf{Mac.}
& \textbf{Wgt.} & \textbf{Mac.} \\
\midrule
\textbf{Zero-shot}
& 62.5 & 41.5
& 42.3 & 26.7
& 40.9 & 31.1
& 58.1 & 40.2
& 56.8 & 43.4
& 52.2 & 38.3 \\
\textbf{IGT}
& 62.5 & 41.5
& 52.7 & 34.0
& 54.4 & 38.1
& 63.8 & 45.4
& 64.5 & 50.5
& 58.5 & 42.8 \\
\textbf{Grammar}
& 62.5 & 41.5
& 55.9 & 41.3
& 60.1 & 47.3
& \textbf{70.7} & \textbf{59.2}
& 69.6 & 54.6
& 69.1 & 56.4 \\
\textbf{Grammar+IGT}
& 62.5 & 41.5
& 50.6 & 40.3
& 52.8 & 41.7
& 67.8 & 53.1
& 67.5 & 52.3
& 65.6 & 54.4 \\
\bottomrule
\end{tabular}
\end{adjustbox}
\caption{Weighted and Macro F1 scores (\%) of \tfc{} under different conditions.}
\label{tab:coding_results}
\end{table*}

\begin{figure*}[htbp]
\centering
\includegraphics[width=\textwidth]{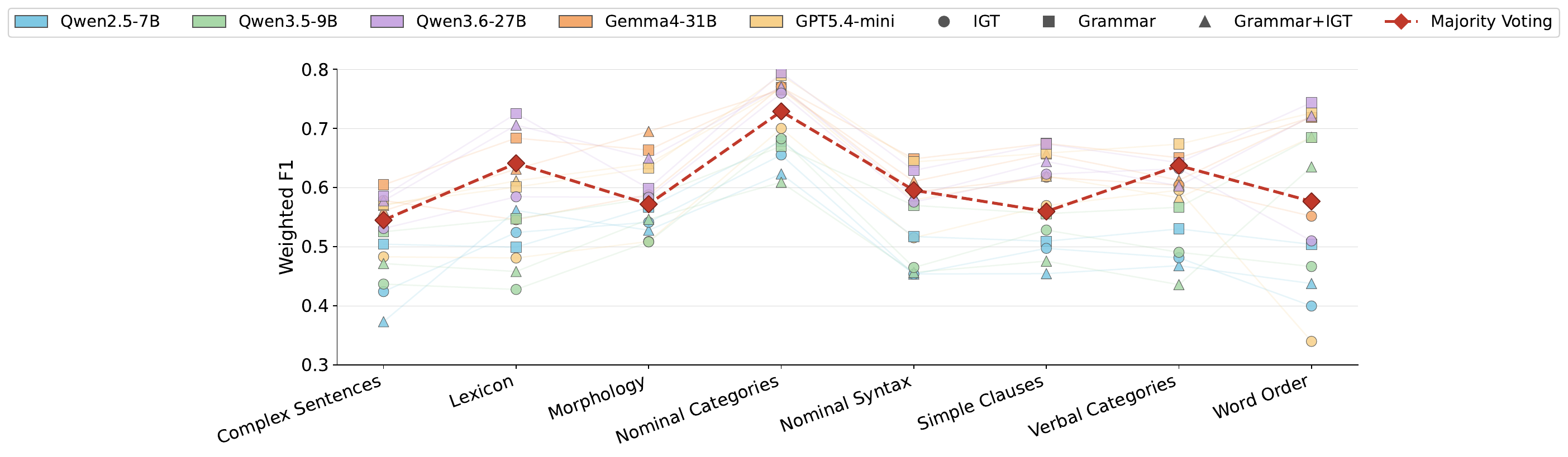}
\caption{Weighted F1 scores of \tfc{} across domains (all settings).}
\label{fig:weighted_f1-domain}
\end{figure*}

\subsection{Metrics} 

For \tfc, we evaluate whether the model's predicted feature values are consistent with the corresponding human annotations in Grambank. Following the Grambank coding scheme, most features are coded with three labels, \{0, 1, ?\}, where 0 and 1 indicate the absence or presence of a feature and ? marks uncertainty or insufficient information. Only 6 features additionally allow multi-valued labels such as 2 or 3 \citep{haynie-etal-2023-grambanks}. Since typological feature labels are often highly imbalanced across languages, we report both \textbf{Weighted F1} and \textbf{Macro F1}. Weighted F1 accounts for label frequency, while Macro F1 gives equal weight to each label.

For \tht, the task is hypothesis-driven and open-ended typological analysis, and typological universals themselves are often statistical tendencies rather than binary truths. Therefore, conventional accuracy, precision, or recall is not well-defined for this task. Instead, we characterize the evidence profile produced by the agent across the language sample. For each language, the agent assigns one of four evidence categories: supporting evidence, counter-evidence, irrelevant evidence, or insufficient evidence. We report the average number of languages assigned to each category across runs. These language-level evidence counts determine whether the hypothesis is ultimately confirmed, refuted, or left inconclusive.

\section{Main Findings}

\subsection{Typological feature coding}
Table~\ref{tab:coding_results} presents the agent's performance on \tfc. Detailed predictions are shown in Appendix \ref{label-prediction-details-all}. 

\paragraph{Performance under different information conditions} Overall, providing access to reference grammars substantially improves typological feature coding performance across nearly all models. Compared with the zero-shot baseline, the Grammar condition consistently yields the strongest gains, particularly for larger models such as \texttt{Qwen3.6-27B} and \texttt{Gemma4-31B}. This suggests that current LLMs are able to effectively utilize explicit grammatical explanations and descriptive prose from reference grammars for typological reasoning. In contrast, the IGT-only condition remains considerably more challenging, although all models still outperform the corresponding zero-shot baseline. This indicates that LLMs can induce non-trivial grammatical generalizations directly from glossed examples, even without access to meta-linguistic explanations. We further examine whether this limitation can be mitigated simply by increasing the ReAct reasoning budget. As shown in Appendix~\ref{app:iteration}, increasing the maximum number of reasoning iterations can consistently improve macro-F1 scores.

Interestingly, the Grammar+IGT condition does not consistently outperform the Grammar-only condition. For most models, adding IGT examples on top of the full grammar leads to only marginal improvements or even slight performance degradation. To assess the reliability of these differences, we additionally
conducted paired language-clustered bootstrap tests with 10,000
resamples, reporting 95\% confidence intervals for the pairwise
differences and Holm-corrected $p$-values in Appendix~\ref{app:statistical_analysis}. The results support our main finding that Grammar+IGT does not significantly improve over Grammar for any of the evaluated models. This suggests that current LLM agents may still rely more heavily on structured grammatical prose than on bottom-up grammatical induction from raw IGT evidence. One possible explanation is that the additional IGT evidence introduces retrieval noise or increases the complexity of evidence integration during the ReAct process. 

\paragraph{Performance across grammatical domains}
We categorize the evaluated features into 8 grammatical domains, following the broad feature-domain organization used in WALS \cite{wals}. This domain-level grouping allows us to evaluate whether model performance varies across different areas of grammar, such as morphology, syntax, and nominal or verbal categories. Details about the feature grouping are shown in Appendix \ref{domain-grouping}. We report our weighted F1 score in Fig. \ref{fig:weighted_f1-domain}. In general, LLMs are more consistent with human coding in \textit{Nominal Categories}, but face more challenges in \textit{Complex Sentences} and \textit{Verbal Categories}. The performance fluctuations across domains are highly aligned with the majority-voting baseline performance, which seems to suggest that prior background knowledge still plays a role.

\begin{figure}[htbp]
\centering
\includegraphics[width=\columnwidth]{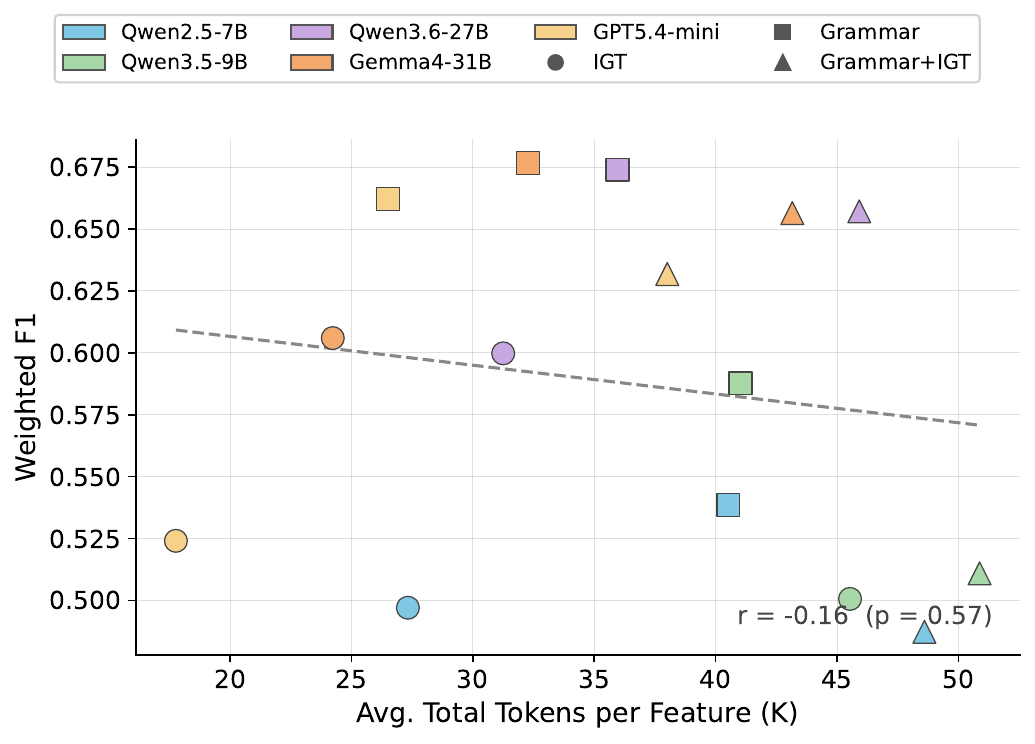}
\caption{Weighted F1 with respect to token usage in \tfc{}. More token usage does not necessarily lead to higher performance.}
\label{fig:weighted_f1-token}
\end{figure}

\begin{table*}[t]
\centering
\small
\resizebox{0.9\textwidth}{!}{%
\setlength{\tabcolsep}{5pt}
\begin{tabular}{ll|rrrr | rrrr}
\toprule
\rowcolor{gray!15}
 & & \multicolumn{4}{c|}{\textbf{Standard}} & \multicolumn{4}{c}{\textbf{Adversarial}} \\
\rowcolor{gray!15}
\textbf{Model} & \textbf{Condition}& \textbf{\# For} & \textbf{\# Against} & \textbf{\# N/A} & \textbf{\# Insuf.} & \textbf{\# For} & \textbf{\# Against} & \textbf{\# N/A} & \textbf{\# Insuf.} \\
\midrule
\multirow{3}{*}{\texttt{Qwen3.6-27B}} & IGT &4.47 & 0.13 & 1.73 & 3.67 & 0.63 & 4.67 & 1.80 & 2.90 \\
 & Grammar& 3.57 & 0.10 & 1.83 & 4.50 & 0.07 & 3.80 & 1.70 & 4.43 \\
 & Grammar+IGT &5.73 & 0.10 & 2.13 & 2.03 & 0.03 & 5.27 & 2.20 & 2.50 \\
\midrule
\multirow{3}{*}{\texttt{Gemma4-31b}} & IGT & 4.77 & 0.60 & 2.37 & 2.27& 1.13 & 4.73 & 2.10 & 2.03 \\
 & Grammar &5.53 & 0.23 & 3.00 & 1.23 & 0.07 & 5.73 & 2.63 & 1.57 \\
 & Grammar+IGT & 6.33 & 0.07 & 2.70 & 0.90 & 0.20 & 5.73 & 2.63 & 1.43 \\
\midrule
\multirow{3}{*}{\texttt{GPT5.4-mini}} & IGT & 1.93 & 0.03 & 1.07 & 6.97& 0.13 & 1.40 & 0.90 & 7.57 \\
 & Grammar & 1.33 & 0.00 & 0.67 & 8.00& 0.03 & 0.87 & 0.73 & 8.37 \\
 & Grammar+IGT & 1.47 & 0.00 & 0.67 & 7.87& 0.07 & 0.83 & 0.63 & 8.47 \\
\bottomrule
\end{tabular}
}
\caption{Average evidence profiles for standard and adversarial universals in TYPOLOGICAL HYPOTHESIS TESTING, averaged over 3 runs and 10 universals. Each row reports the average number of languages assigned to each evidence category (For, Against, N/A, Insuf.) across the 10-language sample; the four counts sum to 10. These counts characterize the cross-linguistic evidence profile rather than prediction accuracy.}
\label{tab:hypotheses_results}
\end{table*}

\paragraph{Token usage} We found that token usage does not necessarily reflect performance, confirming the findings from \newcite{bai2026ai}. As shown in Fig. \ref{fig:weighted_f1-token}, more token usage seems to be associated with lower performance. This is because small models generally consume more tokens than large models, and the IGT-only condition costs less token usage than the other two conditions.

\subsection{Typological hypotheses testing}
Table~\ref{tab:hypotheses_results} illustrates that \agent{} is capable of performing \tht{} across languages independently. Even for adversarial hypotheses, which are deliberately crafted to be false for most languages, \agent{} can still successfully falsify those hypotheses based on available evidence from reference grammars.

Generally speaking, LLMs tend to be less accurate in confirming or refuting hypotheses when only IGTs are given, again highlighting the challenges of inducing grammatical structures from raw data. LLMs are still better at synthesizing information from well-structured English grammatical descriptions, but struggle at reasoning over IGTs from rare or unseen languages.

LLMs also exhibit different behaviors during the agentic analysis. \texttt{GPT5.4-mini} is the most uncertain LLM, as it concludes insufficient evidence for most languages in a batch. In contrast, both \texttt{Gemma4-31B} and \texttt{Qwen3.6-27B} are more willing to draw conclusions from individual languages. We speculate that such differences might arise from the post-training stage.

\section{Error Analysis}

We first condition on incorrect predictions and examine how they are distributed across predicted labels. We check the predictions from \texttt{Qwen3.6-27B} with grammar access. For \tfc{}, as shown in the Table \ref{tab:error_distribution}. The model sometimes treats absent features as present, or misses evidence for features that are actually present. Interestingly, when the gold label is \texttt{?}, incorrect predictions are often forced into categorical labels rather than remaining uncertain.
\begin{table}[t]
\centering
\begin{adjustbox}{width=0.9\columnwidth}
\begin{tabular}{lcccc}
\hline
\textbf{Gold label} & \textbf{Predicted 0} & \textbf{Predicted 1} 
& \textbf{Predicted ?} & \textbf{Other} \\
\hline
0 & --- & 56.0\% & 44.0\% & --- \\
1 & 66.4\% & --- & 31.4\% & 2.2\% \\
? & 67.2\% & 30.8\% & --- & 2.0\% \\
\hline
\end{tabular}
\end{adjustbox}
\caption{Distribution of predicted labels among incorrect predictions, grouped by gold label.}
\label{tab:error_distribution}
\end{table}

To better characterize the failure modes, we conducted a systematic error analysis of prediction--gold mismatches. We pool prediction–gold mismatches from the three models (\texttt{Qwen3.6-27B}, \texttt{Gemma4-31B}, and \texttt{GPT5.4-mini}). From this pooled set, we sample 10 instances for each of the 6 language with 3 information-setting combinations, yielding 180 instances in total. We manually adjudicate the cases against the original grammar texts. The patterns are summarized as below.

\paragraph{1. Exponence-type misclassification.}
The agent retrieves the relevant form but assigns it to the wrong formal category, such as a particle, clitic, affix, auxiliary, or adverb. This is a source of false-positive feature presence. For example, in {Papuan Malay (GB520)}\footnote{\url{https://grambank.clld.org/languages/papu1250}}, the agent codes the feature as present by treating \textit{suda}, \textit{masi}, and \textit{blum} as aspectual auxiliaries. However, the author of the grammar analyzes these forms as adverbs, and the retrieved evidence does not establish that they satisfy the Grambank criterion for auxiliaries.

\paragraph{2. IGT role or construction misinterpretation.}
It means a gloss tag or morpheme is treated as direct evidence for the wrong grammatical role, host, or construction. In {Ch\'{a}cobo (GB092)}\footnote{\url{https://grambank.clld.org/languages/chac1251}},
for instance, the agent misreads the ergative clitic \textit{=wa}, which marks the overt A noun phrase, as a prefix or proclitic indexing A on the verb. This confuses nominal case marking with verbal argument indexing.

\paragraph{3. Over-restrictive feature interpretation.}
The agent adds requirements not present in the coding procedure, such as demanding a dedicated, obligatory, or single-morpheme exponent, and therefore misses distributed or periphrastic marking.
In {Ik (GB083)\footnote{\url{https://grambank.clld.org/languages/ikkk1242}}}, past tense is coded as absent because it is expressed through combinations of aspectual and enclitic morphology rather than a dedicated past marker.

\paragraph{4. Retrieval--terminology mismatch.}
The grammar describes the target phenomenon using terminology different from the Grambank query, so retrieval focuses on related but irrelevant sections. In {Japhug (GB049)}\footnote{\url{https://grambank.clld.org/languages/japh1234}}, the query refers to “object nouns derived from verbs,” whereas the grammar discusses the relevant construction under object participles. The agent instead retrieves sections on nominal derivation and Object–Verb compounds.\footnote{Full trace is in Appendix \ref{grambank-agent-trace-example}.}

\paragraph{5. Evidence-integration failure.}
It means that examples from a non-target construction are treated as feature evidence. In {Chakali (GB318)}\footnote{\url{https://grambank.clld.org/languages/chak1271}}, for example, a bound plural suffix glossed \textit{.PL}  is interpreted as a free plural element, despite the grammar analyzing forms such as ger → gete as noun-class suffixation.


For \tht{}, we analyze the agent trace when testing Greenberg Universal rule (U30\footnote{If the verb has categories of person-number or of gender, it always has tense-mode categories.}). As shown in Appendix \ref{hypo-agent-trace}, the model incorrectly refutes the standard Greenberg universal by treating Ik as a counterexample. This error also appears to come from an overly narrow interpretation of the consequent: the model equates ``tense-mode categories'' with a dedicated or obligatory tense paradigm. The planning step is also suboptimal, since the selected sections for Ik focus on phonology and morphophonology rather than verbal morphology or tense, aspect, and modality. 

These errors highlight the gap between human linguists and \agent{} in interpreting linguistic data meta-linguistically. Human expertise is still necessary in validating the outputs.

\section{Conclusions}
In this study, we propose \agent{}, an LLM agent that can conduct typological research across languages autonomously. Empirical evaluations show that \agent{} can complete \tfc{} and \tht{} with strong performance.
Beyond task performance, our findings highlight the importance of decomposing typological research
into structured tools and steps with traceable evidence.

By automating the otherwise time-consuming work of querying evidence across reference grammars, \agent{} can enable linguists to focus on deeper analysis. Future research can extend this framework to more languages, more modalities (speech and signs), and incorporate interactive feedback from human linguists.

\section*{Limitations}
Our research agent is validated against existing typological databases carefully curated by expert linguists. The values of WALS-like typological databases (Grambank and APiCS) are also not without controversies within the linguistic community from theoretical and empirical grounds \cite[e.g.,][]{davis2014investigate,haspelmath2014descriptive}. A small percentage of coded features in some languages are also found to be inaccurate \cite{haspelmath2014descriptive}. For comparability, such databases reduce linguistic diversity into a limited set of discrete codes, though such discrete codes applied uniformly to languages are needed for cross-linguistic comparison. 

Secondly, while the agent can generate and test new hypotheses, extensive human validation is still needed to validate the scientific value and the correctness of these hypotheses. \agent{} brings a validation asymmetry where providing answers is way easier than validating these answers. Also, \agent{} cannot replace human field linguists who gather linguistic data from fieldwork. Many universal hypotheses in \tht{} still need more linguistic data to validate. 

Thirdly, language is inherently multi-modal. Our current research only focuses on the textual aspects, due to the constraints of available data and LLMs. But the speech and the visual (signing) modalities of language are also essential components of language, which deserve further research. 

Languages are living and fluid, constantly changing with the linguistic communities. \agent{} relies on reference grammars, which are only static snapshots of the living languages. LLM agents cannot replace field linguists who are doing fieldwork to interact with diverse communities. 

\section*{Ethical considerations}
Our research builds on the collective effort from the field linguists and various linguistic communities. Our purpose is not to replace human linguists but to provide tools that facilitate linguistic research. We acknowledge the importance of linguists and fieldwork, not just in collecting language data, but also in building bonds with those linguistic communities.

We only selected reference grammars with permissive licenses like Creative Commons licenses. All reference grammars are cited and will only be used for research purposes. We also implement an opt-out mechanism where the authors and communities can ask to remove their grammars from the agentic system. 

\paragraph{Statements on LLM usage}
We disclose that LLMs had been used to assist with grammar correction and text polishing. LLMs were also used for general coding assistance under human oversight. 

\section*{Acknowledgments}
We thank three anonymous reviewers and the area
chairs for their thoughtful comments. We would also like to extend our gratitude to the field workers and the community members documenting these languages. 

This research was enabled in part through the computational resources provided by Advanced Research Computing at the University of British Columbia and the
Digital Research Alliance of Canada. The research
activities were also supported by the NSERC Discovery Grant and the CFI-JELF Grant awarded to
JZ, and a Canada CIFAR AI Chair Award to FS.

\bibliography{custom}

\appendix

\section{Reference Grammars Details}
\label{sec:all-reference-grammar-info}
Table~\ref{app:reference-grammars} and \ref{app:grammars} shows detailed information about the reference grammar languages and sources we used in the paper.

\begin{table}[ht]
\centering
\resizebox{\columnwidth}{!}{
\begin{tabular}{l l r r}
\toprule
\textbf{Language} & \textbf{Family} & \textbf{Chunk \#} & \textbf{IGT \#} \\
\midrule
Moloko & Afro-Asiatic & 219 & 488 \\
\midrule
Chakali & Atlantic-Congo & 97 & 405 \\
Kam & Atlantic-Congo & 262 & 698 \\
\midrule
Paunaka & Arawakan & 345 & 1604 \\
\midrule
Belep & Austronesian & 303 & 1009 \\
Kagayanen & Austronesian & 296 & 570 \\
Lele & Austronesian & 168 & 494 \\
Nese & Austronesian & 205 & 611 \\
Paluai & Austronesian & 343 & 603 \\
Papuan Malay & Austronesian & 450 & 1155 \\
\midrule
Aguaruna & Chicham & 523 & 512 \\
\midrule
Ulwa & Keram & 354 & 1502 \\
\midrule
Ik & Kuliak & 128 & 103 \\
\midrule
Hup & Naduhup & 425 & 2074 \\
\midrule
Bargam & Nuclear Trans New Guinea & 241 & 823 \\
Mauwake & Nuclear Trans New Guinea & 276 & 1646 \\
Mian & Nuclear Trans New Guinea & 601 & 1013 \\
\midrule
Ch\'{a}cobo & Pano-Tacanan & 452 & 1302 \\
\midrule
Eastern Geshiza & Sino-Tibetan & 580 & 1079 \\
Japhug & Sino-Tibetan & 1232 & 5517 \\
Karbi & Sino-Tibetan & 384 & 681 \\
Yakkha & Sino-Tibetan & 355 & 1312 \\
\midrule
Pite Saami & Uralic & 222 & 179 \\
\midrule
Kalamang & West Bomberai & 337 & 630 \\
\midrule
Komnzo & Yam & 223 & 565 \\
\bottomrule
\end{tabular}
}
\caption{Languages included in the analysis, together with the number of extracted chunks and IGT instances.}
\label{app:reference-grammars}
\end{table}

\begin{table*}[t]
\centering
\small
\setlength{\tabcolsep}{4pt}
\begin{adjustbox}{width=\textwidth}
\begin{tabular}{llp{6.5cm}p{4.5cm}}
\toprule
\textbf{Language} & \textbf{Family} & \textbf{Grambank Source} & \textbf{Our Source*} \\
\midrule

Moloko& Afro-Asiatic & \textcolor{red}{Friesen \& Mamalis. 2004. \textit{The Moloko Verb Phrase}. SIL.} & Friesen. 2017. \textit{A grammar of Moloko}. Language Science Press. \\
\midrule
Paunaka & Arawakan & \textcolor{red}{Danielsen \& Terhart. 2014. Paunaka. In \textit{Oriente}.} & Terhart. 2024. \textit{A grammar of Paunaka}. Language Science Press. \\
\midrule
Chakali & Atlantic-Congo & \textcolor{red}{Brindle. 2010. \textit{Aspects of the Chakali language}. NTNU dissertation.} & Brindle. 2017. \textit{A dictionary and grammatical outline of Chakali}. Language Science Press. \\
Kam & Atlantic-Congo & Lesage. 2020. \textit{A grammar and lexicon of Kam}. INALCO dissertation. & \\
\midrule
Belep & Austronesian & McCracken. 2013. \textit{A grammar of Belep}. Rice University dissertation. & \\
Kagayanen & Austronesian & \textcolor{red}{Pebley \& MacGregor. 1999. \textit{Studies in Kagayanen}. SIL.} & Pebley \& Payne. 2024. \textit{A grammar of Kagayanen}. Language Science Press. \\
Lele& Austronesian & Boettger. 2015. \textit{Topics in the grammar of Lele}. James Cook University dissertation. & \\
Nese& Austronesian & Takau. 2016. \textit{A grammar of Nese}. University of Newcastle dissertation. & \\
Paluai& Austronesian & Schokkin. 2014. \textit{A grammar of Paluai}. James Cook University dissertation. & \\
Papuan Malay& Austronesian & Kluge. 2017. \textit{A grammar of Papuan Malay}. Language Science Press. & \\
\midrule
Aguaruna& Chicham& Overall. 2007. \textit{A Grammar of Aguaruna}. LaTrobe University dissertation. & \\
\midrule
Ulwa (PNG)& Keram& \textcolor{red}{Barlow. 2018. \textit{A grammar of Ulwa}. University of Hawai'i dissertation.} & Barlow. 2023. \textit{A grammar of Ulwa}. Language Science Press. \\
\midrule
Ik& Kuliak & \textcolor{red}{Schrock. 2014. \textit{A grammar of Ik}. Leiden University dissertation.} & Schrock. 2017. \textit{The Ik language}. Language Science Press. \\
\midrule
Hup & Naduhup& Epps. 2008. \textit{A grammar of Hup}. Mouton de Gruyter. & \\
\midrule
Bargam& Nuclear Trans New Guinea & Hepner. 2006. \textit{Bargam Grammar Sketch}. SIL. & \\
Mauwake & Nuclear Trans New Guinea & \textcolor{red}{Berghäll. 2010. \textit{Mauwake reference grammar}. University of Helsinki dissertation.} & Berghäll. 2015. \textit{A grammar of Mauwake}. Language Science Press. \\
Mian& Nuclear Trans New Guinea & Fedden. 2007. \textit{A Grammar of Mian}. University of Melbourne dissertation. & \\
\midrule
Ch\'{a}cobo & Pano-Tacanan & Tallman. 2018. \textit{A Grammar of Ch\'{a}cobo}. University of Texas dissertation. & \\
\midrule
Eastern Geshiza & Sino-Tibetan & Honkasalo. 2019. \textit{A grammar of Eastern Geshiza}. University of Helsinki dissertation. & \\
Japhug& Sino-Tibetan & \textcolor{red}{Jacques. 2004/2017. \textit{Phonologie et Morphologie du Japhug}; Japhug. In \textit{Sino-Tibetan Languages}.} & Jacques. 2021. \textit{A grammar of Japhug}. Language Science Press. \\
Karbi & Sino-Tibetan & \textcolor{red}{Grüßner. 1978. \textit{Arleng Alam Die Sprache der Mikir}. Franz Steiner.} & Philippova, Nailya. 2021. A Grammar of Amri Karbi. Diss. University of Helsinki. \\
Yakkha& Sino-Tibetan & Schackow. 2015. \textit{A grammar of Yakkha}. Language Science Press. & \\
\midrule
Pite Saami& Uralic & Wilbur, J. (p.c.) & \\
\midrule
Kalamang& West Bomberai& \textcolor{red}{Visser. 2016. \textit{A grammar sketch of Kalamang}. University of Oslo MA thesis.} & Visser. 2022. \textit{A grammar of Kalamang}. Language Science Press. \\
\midrule
Komnzo& Yam& \textcolor{red}{Döhler. 2016. \textit{Komnzo}. ANU dissertation.} & Döhler. 2018. \textit{A grammar of Komnzo}. Language Science Press. \\

\bottomrule
\end{tabular}
\end{adjustbox}
\caption{Reference grammars used in this study, grouped by language family. *Where our source differs from the Grambank source (typically a more recent published version of the same grammar), it is listed in the ``Our Source'' column. Grammars highlighted in red were not used in this study.}
\label{app:grammars}
\end{table*}

\section{Greenberg Universals}
\label{greenberg-all-rules-adv}
Table \ref{tab:greenberg-adversarial-rules} shows both the standard Greenberg universals and adversarial counterparts used in the \tht.
\begin{table*}[t]
\centering
\footnotesize
\setlength{\tabcolsep}{4pt}
\renewcommand{\arraystretch}{1.15}
\begin{tabularx}{\textwidth}{l l l X X}
\hline
\textbf{ID} & \textbf{Logic} & \textbf{Domain} & \textbf{Standard universal} & \textbf{Adversarial hypothesis} \\
\hline
U1 & absolute & Word order 
& In declarative sentences with nominal subject and object, the dominant order is almost always one in which the subject precedes the object.
& In declarative sentences with nominal subject and object, the dominant order is almost always one in which the object precedes the subject. \\
\hline
U8 & implication & Information structure
& When a yes-no question is marked by a single morpheme, it is always differentiated from the corresponding declarative by intonation or by a characteristic position of the morpheme at the beginning or end of the sentence.
& When a yes-no question is marked by a single morpheme, it is not differentiated from the corresponding declarative by intonation or by a characteristic position of the morpheme at the beginning or end of the sentence. \\
\hline
U14 & absolute & Word order
& In conditional statements, the conditional clause precedes the conclusion as the normal order in all languages.
& In conditional statements, the conclusion precedes the conditional clause as the normal order in all languages. \\
\hline
U18 & correlation & Noun phrase structure
& When the descriptive adjective precedes the noun, the demonstrative and the numeral, with overwhelmingly more than chance frequency, do likewise.
& When the descriptive adjective precedes the noun, the demonstrative and the numeral, with overwhelmingly more than chance frequency, follow the noun. \\
\hline
U25 & implication & Word order
& If the pronominal object follows the verb, so does the nominal object.
& If the pronominal object follows the verb, the nominal object precedes the verb. \\
\hline
U29 & implication & Morphological complexity
& If a language has inflection, it always has derivation.
& If a language has inflection, it does not have derivation. \\
\hline
U30 & implication & Tense/aspect/modality
& If the verb has categories of person-number or of gender, it always has tense-mode categories.
& If the verb has categories of person-number or of gender, it does not have tense-mode categories. \\
\hline
U34b & implication & Morphological complexity
& No language has a dual unless it has some nonsingular number category.
& A language may have a dual even if it has no other nonsingular number category. \\
\hline
U36 & implication & Agreement
& If a language has the category of gender, it always has the category of number.
& If a language has the category of gender, it does not have the category of number. \\
\hline
U42 & absolute & Morphological complexity
& All languages have pronominal categories involving at least three persons and two numbers.
& All languages have pronominal categories involving fewer than three persons or fewer than two numbers. \\
\hline
\end{tabularx}
\caption{Greenberg universals and their adversarial counterparts used in the hypothesis-testing experiments.}
\label{tab:greenberg-adversarial-rules}
\end{table*}

\section{Automatic Summarization}
\label{app:auto-summary}
Beyond the two evaluated tasks, \textsc{AutoTypologist} supports an automatic summarization mode in which the agent autonomously generates a structured typological profile of a language without any user-supplied queries. 

In this mode, the agent is first presented with the grammar table of contents and aggregate IGT statistics, from which it identifies major typological domains and generates a set of candidate feature questions for each domain. It then executes the full ReAct investigation pipeline for each candidate question, producing an evidence-grounded answer. The resulting feature-level answers are finally aggregated into a structured typological summary covering the language's major grammatical domains, including word order, morphological complexity, tense-aspect-modality, nominal categories, and argument structure, among others.

This mode is intended to support rapid typological documentation of under-described languages, where no specific research question is known in advance. The output is a structured JSON report containing, for each investigated feature, the predicted value, a prose justification grounded in grammar sections and IGT examples, quantitative IGT evidence, and a confidence score. We did not conduct a systematic evaluation of this mode in the present study, as it would require a different evaluation framework from the feature-by-feature comparison against Grambank. We leave a full evaluation of automatic summarization quality to future work.

\section{Label Prediction Details}
\label{label-prediction-details-all}
Tables \ref{tab:perclass_qwen25_7b}, \ref{tab:perclass_qwen35_9b}, \ref{tab:perclass_qwen36_27b}, \ref{tab:perclass_gemma_31b}, and \ref{tab:perclass_gpt54_mini} show detailed prediction labels for various languages, models, and settings.
\begin{table*}[p]
\centering
\tiny
\setlength{\tabcolsep}{3pt}
\begin{adjustbox}{width=\textwidth}
\begin{tabular}{l|rrrrr|rrrrr|rrrrr}
\toprule
\textbf{Language} & \multicolumn{5}{c|}{\textbf{IGT}} & \multicolumn{5}{c|}{\textbf{Grammar}} & \multicolumn{5}{c}{\textbf{Grammar+IGT}} \\
 & \textbf{0} & \textbf{1} & \textbf{2} & \textbf{3} & \textbf{?} & \textbf{0} & \textbf{1} & \textbf{2} & \textbf{3} & \textbf{?} & \textbf{0} & \textbf{1} & \textbf{2} & \textbf{3} & \textbf{?} \\
\midrule
Aguaruna & 0.61 & 0.42 & 0.00 & 0.00 & 0.12 & 0.66 & 0.56 & 0.50 & 1.00 & 0.00 & 0.58 & 0.48 & 0.33 & 1.00 & 0.08 \\
Bargam & 0.58 & 0.36 & 0.00 & \textit{NA} & 0.08 & 0.60 & 0.42 & 0.33 & \textit{NA} & 0.08 & 0.49 & 0.37 & 0.00 & \textit{NA} & 0.05 \\
Belep & 0.65 & 0.35 & 0.40 & \textit{NA} & 0.09 & 0.68 & 0.51 & 0.57 & \textit{NA} & 0.00 & 0.58 & 0.43 & 0.29 & \textit{NA} & 0.08 \\
Chakali & 0.68 & 0.33 & 0.40 & \textit{NA} & 0.00 & 0.74 & 0.44 & 0.57 & \textit{NA} & 0.00 & 0.66 & 0.45 & 0.33 & \textit{NA} & 0.00 \\
Ch\'{a}cobo & 0.67 & 0.56 & 0.00 & 0.00 & 0.00 & 0.60 & 0.57 & 0.00 & 0.67 & 0.14 & 0.57 & 0.54 & 1.00 & 0.67 & 0.08 \\
Eastern Geshiza & 0.66 & 0.50 & 0.00 & \textit{NA} & 0.00 & 0.68 & 0.58 & 0.40 & \textit{NA} & 0.00 & 0.68 & 0.58 & 0.00 & \textit{NA} & 0.00 \\
Hup & 0.62 & 0.43 & 0.00 & \textit{NA} & 0.00 & 0.58 & 0.51 & 0.00 & \textit{NA} & 0.09 & 0.39 & 0.44 & 0.50 & \textit{NA} & 0.00 \\
Ik & 0.72 & 0.26 & 0.57 & 0.00 & 0.00 & 0.65 & 0.46 & 1.00 & 0.00 & 0.00 & 0.70 & 0.47 & 0.91 & 0.00 & 0.22 \\
Japhug & 0.59 & 0.45 & 0.40 & 0.00 & 0.00 & 0.58 & 0.57 & 0.50 & 0.00 & 0.00 & 0.45 & 0.47 & 0.00 & 0.00 & 0.00 \\
Kagayanen & 0.63 & 0.43 & 1.00 & 0.00 & 0.12 & 0.68 & 0.43 & 0.67 & 0.00 & 0.13 & 0.54 & 0.40 & 0.50 & 0.50 & 0.07 \\
Kalamang & 0.53 & 0.18 & 0.50 & \textit{NA} & 0.08 & 0.45 & 0.19 & 0.33 & \textit{NA} & 0.06 & 0.46 & 0.21 & 0.67 & \textit{NA} & 0.06 \\
Kam & 0.75 & 0.40 & 0.67 & \textit{NA} & 0.00 & 0.73 & 0.53 & 0.75 & \textit{NA} & 0.00 & 0.66 & 0.53 & 0.67 & \textit{NA} & 0.00 \\
Karbi & 0.59 & 0.33 & 0.00 & \textit{NA} & 0.04 & 0.51 & 0.38 & 0.50 & \textit{NA} & 0.12 & 0.42 & 0.38 & 0.00 & \textit{NA} & 0.08 \\
Komnzo & 0.65 & 0.46 & \textit{NA} & \textit{NA} & 0.00 & 0.68 & 0.55 & \textit{NA} & \textit{NA} & 0.00 & 0.57 & 0.49 & \textit{NA} & \textit{NA} & 0.29 \\
Lele & 0.64 & 0.36 & 0.50 & 0.00 & 0.05 & 0.68 & 0.48 & 0.75 & 0.00 & 0.07 & 0.65 & 0.48 & 0.67 & 1.00 & 0.06 \\
Mauwake & 0.58 & 0.46 & \textit{NA} & \textit{NA} & 0.00 & 0.65 & 0.62 & \textit{NA} & \textit{NA} & 0.17 & 0.64 & 0.63 & \textit{NA} & \textit{NA} & 0.22 \\
Mian & 0.66 & 0.51 & 0.67 & \textit{NA} & 0.00 & 0.65 & 0.65 & 0.00 & \textit{NA} & 0.00 & 0.59 & 0.60 & 0.00 & \textit{NA} & 0.50 \\
Moloko & 0.53 & 0.35 & 0.80 & \textit{NA} & 0.09 & 0.58 & 0.29 & 0.40 & \textit{NA} & 0.09 & 0.50 & 0.26 & 0.00 & \textit{NA} & 0.04 \\
Nese & 0.73 & 0.38 & 0.40 & 0.00 & 0.00 & 0.75 & 0.48 & 0.00 & 0.00 & 0.00 & 0.67 & 0.35 & 0.57 & 1.00 & 0.00 \\
Paluai & 0.63 & 0.34 & 0.00 & \textit{NA} & 0.25 & 0.66 & 0.48 & 0.57 & \textit{NA} & 0.29 & 0.58 & 0.48 & 0.75 & \textit{NA} & 0.00 \\
Papuan Malay & 0.77 & 0.40 & 0.50 & 0.00 & 0.00 & 0.76 & 0.55 & 0.00 & 1.00 & 0.00 & 0.69 & 0.42 & 0.00 & 0.50 & 0.00 \\
Paunaka & 0.70 & 0.56 & 0.00 & \textit{NA} & 0.00 & 0.68 & 0.51 & 1.00 & \textit{NA} & 0.00 & 0.56 & 0.58 & 1.00 & \textit{NA} & 0.00 \\
Pite Saami & 0.74 & 0.31 & \textit{NA} & \textit{NA} & \textit{NA} & 0.70 & 0.50 & \textit{NA} & \textit{NA} & \textit{NA} & 0.66 & 0.49 & \textit{NA} & \textit{NA} & \textit{NA} \\
Ulwa & 0.72 & 0.43 & 0.33 & 0.00 & 0.00 & 0.73 & 0.49 & 0.75 & 0.00 & 0.00 & 0.61 & 0.41 & 0.57 & 0.00 & 0.22 \\
Yakkha & 0.67 & 0.41 & \textit{NA} & \textit{NA} & 0.22 & 0.71 & 0.53 & \textit{NA} & \textit{NA} & 0.14 & 0.66 & 0.54 & \textit{NA} & \textit{NA} & 0.12 \\
\bottomrule
\end{tabular}
\end{adjustbox}
\caption{Per-class F1 scores for \texttt{Qwen2.5-7B}. \textit{NA} indicates the class does not appear in the gold labels for that language.}
\label{tab:perclass_qwen25_7b}
\end{table*}

\begin{table*}[p]
\centering
\tiny
\setlength{\tabcolsep}{3pt}
\begin{adjustbox}{width=\textwidth}
\begin{tabular}{l|rrrrr|rrrrr|rrrrr}
\toprule
\textbf{Language} & \multicolumn{5}{c|}{\textbf{IGT}} & \multicolumn{5}{c|}{\textbf{Grammar}} & \multicolumn{5}{c}{\textbf{Grammar+IGT}} \\
 & \textbf{0} & \textbf{1} & \textbf{2} & \textbf{3} & \textbf{?} & \textbf{0} & \textbf{1} & \textbf{2} & \textbf{3} & \textbf{?} & \textbf{0} & \textbf{1} & \textbf{2} & \textbf{3} & \textbf{?} \\
\midrule
Aguaruna & 0.66 & 0.51 & 0.00 & 0.00 & 0.11 & 0.69 & 0.61 & 0.67 & 1.00 & 0.29 & 0.50 & 0.58 & 0.67 & 0.50 & 0.43 \\
Bargam & 0.56 & 0.49 & 0.00 & \textit{NA} & 0.27 & 0.62 & 0.62 & 0.67 & \textit{NA} & 0.37 & 0.45 & 0.54 & 0.86 & \textit{NA} & 0.33 \\
Belep & 0.63 & 0.41 & 0.57 & \textit{NA} & 0.20 & 0.65 & 0.51 & 0.33 & \textit{NA} & 0.18 & 0.48 & 0.50 & 0.50 & \textit{NA} & 0.15 \\
Chakali & 0.68 & 0.41 & 0.67 & \textit{NA} & 0.06 & 0.52 & 0.44 & 0.40 & \textit{NA} & 0.09 & 0.36 & 0.51 & 0.00 & \textit{NA} & 0.10 \\
Ch\'{a}cobo & 0.53 & 0.57 & 1.00 & 0.00 & 0.14 & 0.52 & 0.62 & 0.00 & 0.50 & 0.17 & 0.47 & 0.62 & 0.00 & 0.00 & 0.29 \\
Eastern Geshiza & 0.65 & 0.57 & 0.00 & \textit{NA} & 0.07 & 0.58 & 0.62 & 0.50 & \textit{NA} & 0.11 & 0.50 & 0.73 & 0.00 & \textit{NA} & 0.21 \\
Hup & 0.59 & 0.54 & 0.67 & \textit{NA} & 0.06 & 0.61 & 0.64 & 0.50 & \textit{NA} & 0.18 & 0.43 & 0.54 & 0.67 & \textit{NA} & 0.13 \\
Ik & 0.72 & 0.34 & 0.67 & 0.00 & 0.09 & 0.73 & 0.67 & 0.80 & 0.00 & 0.14 & 0.63 & 0.63 & 0.89 & 0.67 & 0.12 \\
Japhug & 0.66 & 0.60 & 0.00 & 0.00 & 0.08 & 0.69 & 0.70 & 0.80 & 0.00 & 0.17 & 0.58 & 0.66 & 0.40 & 0.50 & 0.17 \\
Kagayanen & 0.59 & 0.38 & 0.00 & 0.00 & 0.29 & 0.61 & 0.51 & 0.67 & 0.00 & 0.23 & 0.53 & 0.48 & 0.67 & 0.80 & 0.30 \\
Kalamang & 0.53 & 0.26 & 0.00 & \textit{NA} & 0.20 & 0.59 & 0.29 & 0.33 & \textit{NA} & 0.31 & 0.48 & 0.30 & 0.33 & \textit{NA} & 0.37 \\
Kam & 0.67 & 0.42 & 0.00 & \textit{NA} & 0.00 & 0.71 & 0.58 & 0.80 & \textit{NA} & 0.00 & 0.55 & 0.57 & 0.67 & \textit{NA} & 0.00 \\
Karbi & 0.64 & 0.39 & 0.50 & \textit{NA} & 0.26 & 0.61 & 0.50 & 0.00 & \textit{NA} & 0.36 & 0.44 & 0.43 & 0.00 & \textit{NA} & 0.30 \\
Komnzo & 0.62 & 0.47 & \textit{NA} & \textit{NA} & 0.00 & 0.63 & 0.63 & \textit{NA} & \textit{NA} & 0.04 & 0.61 & 0.65 & \textit{NA} & \textit{NA} & 0.14 \\
Lele & 0.66 & 0.38 & 0.57 & 0.00 & 0.19 & 0.65 & 0.61 & 0.67 & 0.00 & 0.32 & 0.54 & 0.54 & 0.80 & 0.00 & 0.29 \\
Mauwake & 0.59 & 0.54 & \textit{NA} & \textit{NA} & 0.00 & 0.68 & 0.76 & \textit{NA} & \textit{NA} & 0.08 & 0.62 & 0.72 & \textit{NA} & \textit{NA} & 0.08 \\
Mian & 0.57 & 0.43 & 0.67 & \textit{NA} & 0.11 & 0.49 & 0.55 & 0.67 & \textit{NA} & 0.07 & 0.43 & 0.67 & 0.00 & \textit{NA} & 0.00 \\
Moloko & 0.51 & 0.30 & 0.40 & \textit{NA} & 0.21 & 0.50 & 0.46 & 0.00 & \textit{NA} & 0.33 & 0.47 & 0.35 & 0.00 & \textit{NA} & 0.37 \\
Nese & 0.65 & 0.24 & 0.33 & 0.00 & 0.00 & 0.69 & 0.59 & 0.67 & 1.00 & 0.04 & 0.51 & 0.47 & 0.00 & 0.50 & 0.06 \\
Paluai & 0.64 & 0.43 & 0.67 & \textit{NA} & 0.00 & 0.74 & 0.61 & 0.86 & \textit{NA} & 0.12 & 0.62 & 0.54 & 0.33 & \textit{NA} & 0.04 \\
Papuan Malay & 0.82 & 0.45 & 0.50 & 0.00 & 0.00 & 0.76 & 0.73 & 0.50 & 0.67 & 0.11 & 0.76 & 0.70 & 0.50 & 0.50 & 0.11 \\
Paunaka & 0.69 & 0.57 & 0.00 & \textit{NA} & 0.00 & 0.73 & 0.62 & 0.00 & \textit{NA} & 0.00 & 0.59 & 0.55 & 0.00 & \textit{NA} & 0.00 \\
Pite Saami & 0.72 & 0.50 & \textit{NA} & \textit{NA} & \textit{NA} & 0.76 & 0.68 & \textit{NA} & \textit{NA} & \textit{NA} & 0.61 & 0.60 & \textit{NA} & \textit{NA} & \textit{NA} \\
Ulwa & 0.70 & 0.45 & 0.80 & 1.00 & 0.13 & 0.75 & 0.59 & 0.86 & 0.00 & 0.15 & 0.66 & 0.59 & 0.75 & 0.00 & 0.09 \\
Yakkha & 0.67 & 0.56 & \textit{NA} & \textit{NA} & 0.24 & 0.73 & 0.64 & \textit{NA} & \textit{NA} & 0.11 & 0.62 & 0.59 & \textit{NA} & \textit{NA} & 0.10 \\
\bottomrule
\end{tabular}
\end{adjustbox}
\caption{Per-class F1 scores for \texttt{Qwen3.5-9B}. \textit{NA} indicates the class does not appear in the gold labels for that language.}
\label{tab:perclass_qwen35_9b}
\end{table*}

\begin{table*}[p]
\centering
\tiny
\setlength{\tabcolsep}{3pt}
\begin{adjustbox}{width=\textwidth}
\begin{tabular}{l|rrrrr|rrrrr|rrrrr}
\toprule
\textbf{Language} & \multicolumn{5}{c|}{\textbf{IGT}} & \multicolumn{5}{c|}{\textbf{Grammar}} & \multicolumn{5}{c}{\textbf{Grammar+IGT}} \\
 & \textbf{0} & \textbf{1} & \textbf{2} & \textbf{3} & \textbf{?} & \textbf{0} & \textbf{1} & \textbf{2} & \textbf{3} & \textbf{?} & \textbf{0} & \textbf{1} & \textbf{2} & \textbf{3} & \textbf{?} \\
\midrule
Aguaruna & 0.78 & 0.63 & 0.00 & 0.00 & 0.17 & 0.80 & 0.75 & 1.00 & 1.00 & 0.45 & 0.76 & 0.65 & 0.00 & 0.67 & 0.33 \\
Bargam & 0.70 & 0.44 & 0.00 & \textit{NA} & 0.18 & 0.70 & 0.59 & 0.80 & \textit{NA} & 0.30 & 0.67 & 0.60 & 0.00 & \textit{NA} & 0.31 \\
Belep & 0.80 & 0.47 & 0.40 & \textit{NA} & 0.12 & 0.80 & 0.69 & 0.00 & \textit{NA} & 0.13 & 0.80 & 0.72 & 0.67 & \textit{NA} & 0.32 \\
Chakali & 0.86 & 0.55 & 0.00 & \textit{NA} & 0.13 & 0.77 & 0.53 & 0.86 & \textit{NA} & 0.12 & 0.67 & 0.44 & 0.40 & \textit{NA} & 0.13 \\
Ch\'{a}cobo & 0.70 & 0.57 & 0.00 & 0.00 & 0.05 & 0.76 & 0.56 & 1.00 & 0.67 & 0.19 & 0.71 & 0.59 & 0.00 & 0.40 & 0.27 \\
Eastern Geshiza & 0.82 & 0.64 & 0.50 & \textit{NA} & 0.00 & 0.80 & 0.66 & 1.00 & \textit{NA} & 0.23 & 0.74 & 0.69 & 0.50 & \textit{NA} & 0.05 \\
Hup & 0.70 & 0.50 & 0.67 & \textit{NA} & 0.15 & 0.79 & 0.58 & 0.80 & \textit{NA} & 0.14 & 0.69 & 0.56 & 0.50 & \textit{NA} & 0.26 \\
Ik & 0.78 & 0.22 & 0.50 & 0.00 & 0.13 & 0.85 & 0.69 & 1.00 & 0.00 & 0.29 & 0.82 & 0.72 & 0.91 & 0.00 & 0.25 \\
Japhug & 0.76 & 0.67 & 0.00 & 0.00 & 0.10 & 0.79 & 0.72 & 0.00 & 0.67 & 0.15 & 0.75 & 0.69 & 1.00 & 1.00 & 0.11 \\
Kagayanen & 0.78 & 0.51 & 0.00 & 0.67 & 0.23 & 0.80 & 0.55 & 0.67 & 0.00 & 0.00 & 0.74 & 0.50 & 0.67 & 0.50 & 0.23 \\
Kalamang & 0.61 & 0.41 & 0.40 & \textit{NA} & 0.28 & 0.64 & 0.41 & 0.67 & \textit{NA} & 0.17 & 0.63 & 0.34 & 0.57 & \textit{NA} & 0.21 \\
Kam & 0.84 & 0.42 & 0.89 & \textit{NA} & 0.00 & 0.83 & 0.61 & 0.80 & \textit{NA} & 0.00 & 0.81 & 0.66 & 1.00 & \textit{NA} & 0.00 \\
Karbi & 0.75 & 0.52 & 0.50 & \textit{NA} & 0.16 & 0.74 & 0.48 & 0.00 & \textit{NA} & 0.15 & 0.68 & 0.39 & 0.00 & \textit{NA} & 0.25 \\
Komnzo & 0.76 & 0.47 & \textit{NA} & \textit{NA} & 0.14 & 0.87 & 0.63 & \textit{NA} & \textit{NA} & 0.00 & 0.83 & 0.62 & \textit{NA} & \textit{NA} & 0.00 \\
Lele & 0.71 & 0.41 & 0.33 & 0.00 & 0.13 & 0.78 & 0.68 & 1.00 & 0.00 & 0.21 & 0.70 & 0.66 & 0.57 & 0.00 & 0.18 \\
Mauwake & 0.68 & 0.55 & \textit{NA} & \textit{NA} & 0.00 & 0.86 & 0.81 & \textit{NA} & \textit{NA} & 0.00 & 0.82 & 0.78 & \textit{NA} & \textit{NA} & 0.00 \\
Mian & 0.72 & 0.58 & 1.00 & \textit{NA} & 0.00 & 0.92 & 0.82 & 1.00 & \textit{NA} & 0.36 & 0.76 & 0.63 & 1.00 & \textit{NA} & 0.00 \\
Moloko & 0.62 & 0.44 & 0.40 & \textit{NA} & 0.28 & 0.70 & 0.44 & 0.33 & \textit{NA} & 0.21 & 0.67 & 0.53 & 0.57 & \textit{NA} & 0.21 \\
Nese & 0.82 & 0.51 & 0.86 & 0.67 & 0.00 & 0.85 & 0.64 & 0.86 & 0.00 & 0.00 & 0.81 & 0.59 & 0.86 & 1.00 & 0.11 \\
Paluai & 0.78 & 0.42 & 0.86 & \textit{NA} & 0.00 & 0.81 & 0.61 & 0.86 & \textit{NA} & 0.00 & 0.80 & 0.65 & 0.67 & \textit{NA} & 0.11 \\
Papuan Malay & 0.89 & 0.62 & 0.80 & 0.67 & 0.09 & 0.89 & 0.71 & 1.00 & 1.00 & 0.00 & 0.89 & 0.66 & 1.00 & 0.67 & 0.13 \\
Paunaka & 0.77 & 0.55 & 0.00 & \textit{NA} & 0.00 & 0.85 & 0.56 & 0.67 & \textit{NA} & 0.00 & 0.81 & 0.71 & 0.00 & \textit{NA} & 0.50 \\
Pite Saami & 0.83 & 0.59 & \textit{NA} & \textit{NA} & \textit{NA} & 0.87 & 0.74 & \textit{NA} & \textit{NA} & \textit{NA} & 0.86 & 0.73 & \textit{NA} & \textit{NA} & \textit{NA} \\
Ulwa & 0.85 & 0.55 & 0.80 & 1.00 & 0.08 & 0.87 & 0.65 & 1.00 & 0.00 & 0.14 & 0.83 & 0.63 & 0.80 & 0.00 & 0.11 \\
Yakkha & 0.80 & 0.57 & \textit{NA} & \textit{NA} & 0.17 & 0.86 & 0.67 & \textit{NA} & \textit{NA} & 0.27 & 0.84 & 0.73 & \textit{NA} & \textit{NA} & 0.19 \\
\bottomrule
\end{tabular}
\end{adjustbox}
\caption{Per-class F1 scores for \texttt{Qwen3.6-27B}. \textit{NA} indicates the class does not appear in the gold labels for that language.}
\label{tab:perclass_qwen36_27b}
\end{table*}

\begin{table*}[p]
\centering
\tiny
\setlength{\tabcolsep}{3pt}
\begin{adjustbox}{width=\textwidth}
\begin{tabular}{l|rrrrr|rrrrr|rrrrr}
\toprule
\textbf{Language} & \multicolumn{5}{c|}{\textbf{IGT}} & \multicolumn{5}{c|}{\textbf{Grammar}} & \multicolumn{5}{c}{\textbf{Grammar+IGT}} \\
 & \textbf{0} & \textbf{1} & \textbf{2} & \textbf{3} & \textbf{?} & \textbf{0} & \textbf{1} & \textbf{2} & \textbf{3} & \textbf{?} & \textbf{0} & \textbf{1} & \textbf{2} & \textbf{3} & \textbf{?} \\
\midrule
Aguaruna & 0.76 & 0.62 & 1.00 & 0.00 & 0.06 & 0.80 & 0.70 & 1.00 & 0.67 & 0.41 & 0.78 & 0.74 & 0.00 & 0.50 & 0.22 \\
Bargam & 0.73 & 0.51 & 0.86 & \textit{NA} & 0.26 & 0.74 & 0.65 & 1.00 & \textit{NA} & 0.27 & 0.70 & 0.49 & 1.00 & \textit{NA} & 0.15 \\
Belep & 0.76 & 0.49 & 1.00 & \textit{NA} & 0.27 & 0.83 & 0.67 & 0.86 & \textit{NA} & 0.17 & 0.77 & 0.71 & 0.80 & \textit{NA} & 0.46 \\
Chakali & 0.80 & 0.42 & 0.80 & \textit{NA} & 0.09 & 0.82 & 0.64 & 0.80 & \textit{NA} & 0.20 & 0.76 & 0.62 & 1.00 & \textit{NA} & 0.08 \\
Ch\'{a}cobo & 0.69 & 0.65 & 0.00 & 0.00 & 0.24 & 0.71 & 0.69 & 0.00 & 0.67 & 0.17 & 0.67 & 0.71 & 0.00 & 0.57 & 0.23 \\
Eastern Geshiza & 0.82 & 0.66 & 0.80 & \textit{NA} & 0.00 & 0.86 & 0.77 & 0.80 & \textit{NA} & 0.00 & 0.79 & 0.73 & 0.80 & \textit{NA} & 0.00 \\
Hup & 0.69 & 0.51 & 0.50 & \textit{NA} & 0.19 & 0.74 & 0.68 & 1.00 & \textit{NA} & 0.00 & 0.66 & 0.70 & 1.00 & \textit{NA} & 0.11 \\
Ik & 0.74 & 0.41 & 0.50 & 0.00 & 0.17 & 0.85 & 0.73 & 0.91 & 0.00 & 0.10 & 0.82 & 0.68 & 0.89 & 0.00 & 0.14 \\
Japhug & 0.73 & 0.63 & 1.00 & 0.00 & 0.12 & 0.78 & 0.74 & 0.50 & 0.00 & 0.00 & 0.72 & 0.71 & 0.50 & 0.00 & 0.14 \\
Kagayanen & 0.79 & 0.62 & 0.00 & 0.00 & 0.37 & 0.76 & 0.58 & 0.00 & 0.50 & 0.08 & 0.76 & 0.50 & 0.00 & 0.00 & 0.07 \\
Kalamang & 0.58 & 0.42 & 0.40 & \textit{NA} & 0.23 & 0.64 & 0.42 & 0.80 & \textit{NA} & 0.11 & 0.63 & 0.35 & 0.40 & \textit{NA} & 0.11 \\
Kam & 0.82 & 0.53 & 0.86 & \textit{NA} & 0.00 & 0.82 & 0.65 & 0.80 & \textit{NA} & 0.00 & 0.78 & 0.67 & 0.75 & \textit{NA} & 0.00 \\
Karbi & 0.67 & 0.50 & 0.50 & \textit{NA} & 0.13 & 0.69 & 0.56 & 0.00 & \textit{NA} & 0.00 & 0.62 & 0.50 & 0.50 & \textit{NA} & 0.11 \\
Komnzo & 0.72 & 0.55 & \textit{NA} & \textit{NA} & 0.08 & 0.81 & 0.68 & \textit{NA} & \textit{NA} & 0.00 & 0.80 & 0.71 & \textit{NA} & \textit{NA} & 0.11 \\
Lele & 0.74 & 0.51 & 0.75 & 0.00 & 0.18 & 0.79 & 0.70 & 0.75 & 0.00 & 0.22 & 0.80 & 0.71 & 0.89 & 0.00 & 0.31 \\
Mauwake & 0.75 & 0.60 & \textit{NA} & \textit{NA} & 0.08 & 0.83 & 0.85 & \textit{NA} & \textit{NA} & 0.00 & 0.77 & 0.81 & \textit{NA} & \textit{NA} & 0.00 \\
Mian & 0.67 & 0.59 & 1.00 & \textit{NA} & 0.00 & 0.74 & 0.79 & 1.00 & \textit{NA} & 0.22 & 0.75 & 0.75 & 1.00 & \textit{NA} & 0.00 \\
Moloko & 0.65 & 0.49 & 0.57 & \textit{NA} & 0.24 & 0.63 & 0.45 & 0.40 & \textit{NA} & 0.17 & 0.60 & 0.44 & 0.40 & \textit{NA} & 0.02 \\
Nese & 0.82 & 0.47 & 0.80 & 0.00 & 0.08 & 0.85 & 0.68 & 0.33 & 0.00 & 0.00 & 0.82 & 0.70 & 1.00 & 0.00 & 0.00 \\
Paluai & 0.79 & 0.44 & 0.86 & \textit{NA} & 0.00 & 0.83 & 0.67 & 0.67 & \textit{NA} & 0.00 & 0.87 & 0.77 & 0.67 & \textit{NA} & 0.00 \\
Papuan Malay & 0.92 & 0.64 & 1.00 & 0.00 & 0.14 & 0.87 & 0.68 & 0.67 & 0.00 & 0.00 & 0.90 & 0.73 & 0.80 & 0.00 & 0.12 \\
Paunaka & 0.78 & 0.64 & 1.00 & \textit{NA} & 0.20 & 0.79 & 0.59 & 0.00 & \textit{NA} & 0.25 & 0.73 & 0.63 & 0.67 & \textit{NA} & 0.00 \\
Pite Saami & 0.84 & 0.58 & \textit{NA} & \textit{NA} & \textit{NA} & 0.86 & 0.72 & \textit{NA} & \textit{NA} & \textit{NA} & 0.87 & 0.76 & \textit{NA} & \textit{NA} & \textit{NA} \\
Ulwa & 0.85 & 0.62 & 0.67 & 0.00 & 0.15 & 0.84 & 0.69 & 0.86 & 0.00 & 0.12 & 0.77 & 0.64 & 0.80 & 0.00 & 0.00 \\
Yakkha & 0.77 & 0.61 & \textit{NA} & \textit{NA} & 0.14 & 0.83 & 0.70 & \textit{NA} & \textit{NA} & 0.40 & 0.83 & 0.74 & \textit{NA} & \textit{NA} & 0.14 \\
\bottomrule
\end{tabular}
\end{adjustbox}
\caption{Per-class F1 scores for \texttt{Gemma4-31B}. \textit{NA} indicates the class does not appear in the gold labels for that language.}
\label{tab:perclass_gemma_31b}
\end{table*}

\begin{table*}[p]
\centering
\tiny
\setlength{\tabcolsep}{3pt}
\begin{adjustbox}{width=\textwidth}
\begin{tabular}{l|rrrrr|rrrrr|rrrrr}
\toprule
\textbf{Language} & \multicolumn{5}{c|}{\textbf{IGT}} & \multicolumn{5}{c|}{\textbf{Grammar}} & \multicolumn{5}{c}{\textbf{Grammar+IGT}} \\
 & \textbf{0} & \textbf{1} & \textbf{2} & \textbf{3} & \textbf{?} & \textbf{0} & \textbf{1} & \textbf{2} & \textbf{3} & \textbf{?} & \textbf{0} & \textbf{1} & \textbf{2} & \textbf{3} & \textbf{?} \\
\midrule
Aguaruna & 0.77 & 0.58 & 0.67 & 0.00 & 0.41 & 0.75 & 0.67 & 1.00 & 0.67 & 0.33 & 0.70 & 0.67 & 1.00 & 0.50 & 0.30 \\
Bargam & 0.63 & 0.41 & 0.86 & \textit{NA} & 0.43 & 0.76 & 0.66 & 0.57 & \textit{NA} & 0.13 & 0.64 & 0.60 & 0.86 & \textit{NA} & 0.31 \\
Belep & 0.64 & 0.41 & 0.67 & \textit{NA} & 0.17 & 0.79 & 0.70 & 0.75 & \textit{NA} & 0.32 & 0.76 & 0.70 & 0.86 & \textit{NA} & 0.48 \\
Chakali & 0.76 & 0.42 & 0.86 & \textit{NA} & 0.10 & 0.77 & 0.55 & 1.00 & \textit{NA} & 0.13 & 0.69 & 0.44 & 0.86 & \textit{NA} & 0.12 \\
Ch\'{a}cobo & 0.65 & 0.50 & 0.00 & 0.00 & 0.13 & 0.68 & 0.63 & 1.00 & 0.67 & 0.25 & 0.72 & 0.68 & 0.00 & 0.33 & 0.30 \\
Eastern Geshiza & 0.70 & 0.52 & 0.67 & \textit{NA} & 0.07 & 0.85 & 0.73 & 0.80 & \textit{NA} & 0.16 & 0.72 & 0.69 & 0.80 & \textit{NA} & 0.18 \\
Hup & 0.71 & 0.50 & 0.00 & \textit{NA} & 0.21 & 0.77 & 0.69 & 1.00 & \textit{NA} & 0.14 & 0.67 & 0.62 & 0.40 & \textit{NA} & 0.34 \\
Ik & 0.68 & 0.23 & 0.80 & 0.00 & 0.11 & 0.85 & 0.69 & 0.89 & 0.00 & 0.19 & 0.81 & 0.72 & 0.83 & 0.00 & 0.14 \\
Japhug & 0.66 & 0.54 & 0.40 & 0.00 & 0.12 & 0.80 & 0.77 & 0.67 & 0.00 & 0.15 & 0.64 & 0.67 & 0.00 & 0.50 & 0.18 \\
Kagayanen & 0.65 & 0.35 & 0.40 & 0.00 & 0.22 & 0.74 & 0.48 & 0.67 & 0.00 & 0.04 & 0.71 & 0.53 & 0.67 & 0.67 & 0.08 \\
Kalamang & 0.61 & 0.29 & 0.57 & \textit{NA} & 0.36 & 0.64 & 0.34 & 0.57 & \textit{NA} & 0.14 & 0.63 & 0.34 & 0.67 & \textit{NA} & 0.24 \\
Kam & 0.73 & 0.39 & 0.75 & \textit{NA} & 0.03 & 0.83 & 0.64 & 0.89 & \textit{NA} & 0.00 & 0.80 & 0.66 & 0.86 & \textit{NA} & 0.00 \\
Karbi & 0.68 & 0.44 & 0.00 & \textit{NA} & 0.31 & 0.71 & 0.53 & 0.00 & \textit{NA} & 0.33 & 0.60 & 0.49 & 0.67 & \textit{NA} & 0.33 \\
Komnzo & 0.63 & 0.44 & \textit{NA} & \textit{NA} & 0.09 & 0.81 & 0.65 & \textit{NA} & \textit{NA} & 0.00 & 0.79 & 0.65 & \textit{NA} & \textit{NA} & 0.18 \\
Lele & 0.72 & 0.37 & 0.57 & 0.00 & 0.30 & 0.78 & 0.66 & 0.89 & 0.00 & 0.13 & 0.75 & 0.67 & 0.89 & 0.00 & 0.33 \\
Mauwake & 0.59 & 0.45 & \textit{NA} & \textit{NA} & 0.00 & 0.80 & 0.77 & \textit{NA} & \textit{NA} & 0.00 & 0.78 & 0.80 & \textit{NA} & \textit{NA} & 0.13 \\
Mian & 0.68 & 0.68 & 0.67 & \textit{NA} & 0.00 & 0.80 & 0.73 & 1.00 & \textit{NA} & 0.00 & 0.70 & 0.68 & 1.00 & \textit{NA} & 0.00 \\
Moloko & 0.58 & 0.43 & 0.57 & \textit{NA} & 0.39 & 0.69 & 0.52 & 0.57 & \textit{NA} & 0.10 & 0.56 & 0.45 & 0.57 & \textit{NA} & 0.09 \\
Nese & 0.74 & 0.27 & 0.86 & 0.00 & 0.06 & 0.83 & 0.67 & 1.00 & 0.00 & 0.29 & 0.78 & 0.62 & 0.80 & 0.00 & 0.07 \\
Paluai & 0.68 & 0.37 & 0.57 & \textit{NA} & 0.03 & 0.80 & 0.64 & 0.67 & \textit{NA} & 0.11 & 0.77 & 0.63 & 1.00 & \textit{NA} & 0.08 \\
Papuan Malay & 0.84 & 0.51 & 1.00 & 0.00 & 0.00 & 0.89 & 0.79 & 0.80 & 0.67 & 0.27 & 0.88 & 0.68 & 0.80 & 0.67 & 0.12 \\
Paunaka & 0.74 & 0.52 & 0.00 & \textit{NA} & 0.11 & 0.76 & 0.53 & 0.00 & \textit{NA} & 0.14 & 0.69 & 0.60 & 0.67 & \textit{NA} & 0.22 \\
Pite Saami & 0.77 & 0.58 & \textit{NA} & \textit{NA} & \textit{NA} & 0.90 & 0.76 & \textit{NA} & \textit{NA} & \textit{NA} & 0.84 & 0.74 & \textit{NA} & \textit{NA} & \textit{NA} \\
Ulwa & 0.73 & 0.38 & 0.67 & 0.00 & 0.12 & 0.87 & 0.63 & 0.86 & 0.00 & 0.09 & 0.79 & 0.62 & 0.00 & 0.00 & 0.25 \\
Yakkha & 0.70 & 0.55 & \textit{NA} & \textit{NA} & 0.00 & 0.81 & 0.70 & \textit{NA} & \textit{NA} & 0.21 & 0.75 & 0.63 & \textit{NA} & \textit{NA} & 0.20 \\
\bottomrule
\end{tabular}
\end{adjustbox}
\caption{Per-class F1 scores for \texttt{GPT5.4-mini}. \textit{NA} indicates the class does not appear in the gold labels for that language.}
\label{tab:perclass_gpt54_mini}
\end{table*}

\section{Majority Voting Labels}
\label{majority-voting-details}
Table \ref{tab:majority_labels} shows details about the majority voting labels for all Grambank features.

\begin{table*}[p]
\centering
\small
\setlength{\tabcolsep}{3pt}
\begin{tabular}{llrr||llrr||llrr}
\toprule
\textbf{ID} & \textbf{Lbl} & \textbf{Count} & \textbf{Pr.} & \textbf{ID} & \textbf{Lbl} & \textbf{Count} & \textbf{Pr.} & \textbf{ID} & \textbf{Lbl} & \textbf{Count} & \textbf{Pr.} \\
\midrule
GB020 & 0 & 1374/2403 & 0.57 & GB113 & 1 & 1316/2386 & 0.55 & GB263 & 0 & 1065/2233 & 0.48 \\
GB021 & 0 & 1920/2425 & 0.79 & GB114 & 0 & 1101/2397 & 0.46 & GB264 & 0 & 1561/2231 & 0.70 \\
GB022 & 0 & 1644/2370 & 0.69 & GB115 & 1 & 1135/2405 & 0.47 & GB265 & ? & 906/2189 & 0.41 \\
GB023 & 0 & 1687/2376 & 0.71 & GB116 & 0 & 1846/2310 & 0.80 & GB266 & ? & 934/2186 & 0.43 \\
GB024 & 2 & 1087/2415 & 0.45 & GB117 & 1 & 1152/2407 & 0.48 & GB270 & 0 & 1039/2180 & 0.48 \\
GB025 & 2 & 1066/2425 & 0.44 & GB118 & 1 & 840/2410 & 0.35 & GB273 & ? & 950/2189 & 0.43 \\
GB026 & 0 & 1660/2340 & 0.71 & GB119 & 0 & 1356/2195 & 0.62 & GB275 & 0 & 1051/2188 & 0.48 \\
GB027 & 1 & 1065/2421 & 0.44 & GB120 & 0 & 1301/2200 & 0.59 & GB276 & ? & 888/2188 & 0.41 \\
GB028 & 1 & 1243/2451 & 0.51 & GB121 & 0 & 1522/2216 & 0.69 & GB285 & 0 & 1689/2198 & 0.77 \\
GB030 & 0 & 1804/2438 & 0.74 & GB122 & 0 & 1103/2405 & 0.46 & GB286 & 0 & 1460/2247 & 0.65 \\
GB031 & 0 & 1718/2394 & 0.72 & GB123 & ? & 1124/2388 & 0.47 & GB291 & 0 & 1635/2158 & 0.76 \\
GB035 & 0 & 1111/2428 & 0.46 & GB124 & 0 & 1390/2384 & 0.58 & GB296 & ? & 1164/2137 & 0.54 \\
GB036 & 0 & 2005/2395 & 0.84 & GB126 & 1 & 1191/2366 & 0.50 & GB297 & 0 & 1513/2141 & 0.71 \\
GB037 & 0 & 1922/2389 & 0.80 & GB127 & 0 & 1109/2339 & 0.47 & GB298 & 0 & 1795/2189 & 0.82 \\
GB038 & 0 & 2080/2365 & 0.88 & GB129 & 0 & 2061/2361 & 0.87 & GB299 & 1 & 1342/2194 & 0.61 \\
GB039 & 0 & 1625/2366 & 0.69 & GB130 & 1 & 1827/2415 & 0.76 & GB300 & 0 & 1152/2126 & 0.54 \\
GB041 & 0 & 1600/2352 & 0.68 & GB131 & 0 & 1878/2448 & 0.77 & GB301 & ? & 997/2127 & 0.47 \\
GB042 & 0 & 1916/2442 & 0.78 & GB132 & 0 & 1216/2433 & 0.50 & GB302 & 0 & 1439/2129 & 0.68 \\
GB043 & 0 & 2117/2398 & 0.88 & GB133 & 0 & 1322/2446 & 0.54 & GB303 & 0 & 1524/2129 & 0.72 \\
GB044 & 1 & 1282/2398 & 0.53 & GB134 & 1 & 1644/2361 & 0.70 & GB304 & 0 & 797/2131 & 0.37 \\
GB046 & 0 & 1148/2328 & 0.49 & GB135 & 1 & 1422/2360 & 0.60 & GB305 & 0 & 855/2187 & 0.39 \\
GB047 & 1 & 1182/2418 & 0.49 & GB136 & 1 & 1245/2425 & 0.51 & GB306 & 0 & 1170/2131 & 0.55 \\
GB048 & 1 & 1090/2417 & 0.45 & GB137 & 0 & 1364/2382 & 0.57 & GB309 & 0 & 1325/2172 & 0.61 \\
GB049 & 1 & 1022/2414 & 0.42 & GB138 & 0 & 1299/2377 & 0.55 & GB312 & 1 & 1344/2133 & 0.63 \\
GB051 & 0 & 1796/2383 & 0.75 & GB139 & 1 & 1517/2442 & 0.62 & GB313 & 0 & 1233/2131 & 0.58 \\
GB052 & 0 & 2073/2428 & 0.85 & GB140 & 0 & 796/2376 & 0.34 & GB314 & 0 & 1721/2130 & 0.81 \\
GB053 & 0 & 1667/2381 & 0.70 & GB146 & ? & 983/2351 & 0.42 & GB315 & 0 & 1690/2131 & 0.79 \\
GB054 & 0 & 2036/2408 & 0.85 & GB147 & 0 & 1114/2431 & 0.46 & GB316 & 0 & 1908/2131 & 0.90 \\
GB057 & 0 & 1655/2435 & 0.68 & GB148 & 0 & 1649/2407 & 0.69 & GB317 & 0 & 1929/2131 & 0.91 \\
GB058 & 0 & 2013/2428 & 0.83 & GB149 & 0 & 1869/2352 & 0.79 & GB318 & 0 & 1676/2181 & 0.77 \\
GB059 & 0 & 1070/2387 & 0.45 & GB150 & 0 & 1075/2395 & 0.45 & GB319 & 0 & 2016/2132 & 0.95 \\
GB065 & 1 & 1040/2420 & 0.43 & GB151 & 0 & 1485/2401 & 0.62 & GB320 & 0 & 1993/2132 & 0.93 \\
GB068 & 0 & 1002/2436 & 0.41 & GB152 & 0 & 1126/2405 & 0.47 & GB321 & 0 & 1745/2132 & 0.82 \\
GB069 & 0 & 1454/2381 & 0.61 & GB155 & 1 & 1494/2425 & 0.62 & GB322 & 0 & 1204/2185 & 0.55 \\
GB070 & 0 & 1485/2377 & 0.62 & GB156 & 0 & 1740/2370 & 0.73 & GB323 & 0 & 1000/2186 & 0.46 \\
GB071 & 0 & 1209/2373 & 0.51 & GB158 & 1 & 1387/2385 & 0.58 & GB324 & 0 & 1622/2131 & 0.76 \\
GB072 & 0 & 1184/2375 & 0.50 & GB159 & 1 & 866/2371 & 0.37 & GB325 & 0 & 1099/2128 & 0.52 \\
GB073 & 0 & 1143/2363 & 0.48 & GB160 & 1 & 1153/2377 & 0.49 & GB326 & 1 & 941/2182 & 0.43 \\
GB074 & 1 & 1245/2443 & 0.51 & GB165 & 0 & 2125/2214 & 0.96 & GB327 & 1 & 1429/2188 & 0.65 \\
GB075 & 0 & 1098/2390 & 0.46 & GB166 & 0 & 2090/2213 & 0.94 & GB328 & 0 & 1106/2187 & 0.51 \\
GB079 & 1 & 1468/2313 & 0.63 & GB167 & 0 & 1488/2195 & 0.68 & GB329 & 0 & 1234/2184 & 0.57 \\
GB080 & 1 & 1894/2365 & 0.80 & GB170 & 0 & 1667/2264 & 0.74 & GB330 & 0 & 1235/2172 & 0.57 \\
GB081 & 0 & 1828/2359 & 0.77 & GB171 & 0 & 1622/2243 & 0.72 & GB331 & 0 & 1263/2172 & 0.58 \\
GB082 & 0 & 1482/2392 & 0.62 & GB172 & 0 & 1935/2228 & 0.87 & GB333 & 1 & 1234/2189 & 0.56 \\
GB083 & 1 & 1189/2393 & 0.50 & GB177 & 0 & 1742/2196 & 0.79 & GB334 & 0 & 1394/2188 & 0.64 \\
GB084 & 0 & 1244/2402 & 0.52 & GB184 & 0 & 1199/2202 & 0.54 & GB335 & 0 & 1400/2132 & 0.66 \\
GB086 & 1 & 1424/2339 & 0.61 & GB185 & 0 & 1103/2211 & 0.50 & GB336 & 0 & 1650/2130 & 0.77 \\
GB089 & 0 & 1379/2391 & 0.58 & GB186 & 0 & 1686/2195 & 0.77 & GB400 & 0 & 1400/1893 & 0.74 \\
GB090 & 0 & 1352/2396 & 0.56 & GB187 & 0 & 813/2194 & 0.37 & GB401 & ? & 1050/1891 & 0.56 \\
GB091 & 0 & 1464/2404 & 0.61 & GB188 & 0 & 1087/2183 & 0.50 & GB402 & 0 & 1065/1889 & 0.56 \\
GB092 & 0 & 1430/2405 & 0.59 & GB192 & 0 & 1953/2228 & 0.88 & GB403 & 0 & 965/1889 & 0.51 \\
GB093 & 0 & 1520/2410 & 0.63 & GB193 & 2 & 1237/2262 & 0.55 & GB408 & 0 & 978/1909 & 0.51 \\
GB094 & 0 & 1790/2407 & 0.74 & GB196 & 0 & 2051/2215 & 0.93 & GB409 & 0 & 1379/1919 & 0.72 \\
GB095 & 0 & 1798/2362 & 0.76 & GB197 & 0 & 2143/2216 & 0.97 & GB410 & 1 & 1218/1890 & 0.64 \\
GB096 & 0 & 1799/2357 & 0.76 & GB198 & 0 & 1664/2208 & 0.75 & GB415 & 0 & 1497/1955 & 0.77 \\
GB098 & 0 & 1679/2359 & 0.71 & GB203 & 2 & 795/2194 & 0.36 & GB421 & ? & 778/1889 & 0.41 \\
GB099 & 0 & 1796/2351 & 0.76 & GB204 & ? & 1056/2183 & 0.48 & GB422 & 0 & 984/1889 & 0.52 \\
GB103 & 0 & 1364/2413 & 0.57 & GB250 & 0 & 855/2174 & 0.39 & GB430 & 0 & 1714/1961 & 0.87 \\
GB104 & 0 & 1676/2398 & 0.70 & GB252 & 0 & 1105/2171 & 0.51 & GB431 & 0 & 1475/1961 & 0.75 \\
GB105 & 1 & 1075/2394 & 0.45 & GB253 & 0 & 1179/2170 & 0.54 & GB432 & 0 & 1335/1960 & 0.68 \\
GB107 & 0 & 1225/2416 & 0.51 & GB254 & 0 & 865/2171 & 0.40 & GB433 & 0 & 1195/1963 & 0.61 \\
GB108 & 0 & 1306/2391 & 0.55 & GB256 & 0 & 1114/2170 & 0.51 & GB519 & 0 & 956/1898 & 0.50 \\
GB109 & 0 & 1648/2390 & 0.69 & GB257 & 1 & 835/2254 & 0.37 & GB520 & 0 & 1027/1904 & 0.54 \\
GB110 & 0 & 1692/2404 & 0.70 & GB260 & 0 & 1706/2251 & 0.76 & GB521 & 0 & 1282/1925 & 0.67 \\
GB111 & 0 & 1318/2357 & 0.56 & GB262 & 0 & 1513/2234 & 0.68 & GB522 & 1 & 1135/1893 & 0.60 \\
\bottomrule
\end{tabular}
\caption{Majority voting labels for all Grambank features used in this study. Count shows the number of languages coded with the majority label out of total valid codings. Pr.\ indicates the proportion.}
\label{tab:majority_labels}
\end{table*}

\section{Grammar Chunk and IGT entry example}
\label{Grammar Chunk and IGT entry example}
Figures \ref{fig:chunk-example} and \ref{fig:igt-example} show the example of grammar chunk and IGT entry, respectively.

\begin{figure*}[t]
\centering
\begin{tabular}{lp{12cm}}
\hline
\multicolumn{2}{c}{\textbf{Example of a structured grammar chunk}} \\
\hline
language & Kagayanen \\
chunk\_id & chunk\_0087\_p0 \\
chapter & Verb structure and inflection \\
section & Overall verb structure \\
subsection & \\
subsubsection & \\
text & A verbal predicate in Kagayanen consists minimally of a stem, plus one and only one inflectional affix. Verbal predicate = Infl-Stem. A stem consists minimally of a root, and may contain one or more stem-forming processes: Stem = (SF)-Root-(SF) ... \\
summary & In Kagayanen, a verbal predicate minimally consists of a stem plus exactly one inflectional affix, with optional adverbial elements appearing before or after this core structure. The stem itself minimally contains a root and may also include one or more stem-forming processes. \\
\hline
\end{tabular}
\caption{Example of a structured grammar chunk. Each chunk has a unique identifier and stores the original text together with hierarchical metadata and a short summary.}
\label{fig:chunk-example}
\end{figure*}

\begin{figure*}[t]
\centering
\begin{tabular}{lp{12cm}}
\hline
\multicolumn{2}{c}{\textbf{Example of a structured IGT entry}} \\
\hline
language & Kagayanen \\
example\_id & ex:thecabinet \\
source & Palimpyuan a din ta aparador. \\
morpheme & Pa-limpyo-an a din ta aparador. \\
gloss & T.R-clean-APL 1S.ABS 3S.ERG NABS cabinet \\
translation & S/he cleaned the cabinet for me. (Meaning I will not have to clean it.) \\
gloss\_tags & [T.R, clean, APL], [1S.ABS], [3S.ERG], [NABS], [cabinet] \\
chapter & Voice \\
section & The Kagayanen choir \\
subsection & Applicative constructions \\
subsubsection & \\
\hline
\end{tabular}
\caption{Example of a structured IGT entry. Each IGT example has a unique identifier and stores the source sentence, morpheme-segmented line, gloss line, free translation, token-level gloss tags, and the corresponding structural metadata from the original grammar.}
\label{fig:igt-example}
\end{figure*}

\section{Hyperparameters}\label{app: hyper}
Table \ref{tab:hyperparameters} describes different types of system hyperparameters.

\begin{table*}[t]
\centering
\begin{tabular}{lll}
\hline
\multicolumn{3}{c}{\textbf{System Hyperparameters}} \\
\hline
\textbf{Category} & \textbf{Parameter} & \textbf{Value} \\
\hline
\multirow{5}{*}{LLM}
& Temperature & 0.1 \\
& Max tokens& 32768 \\
& Force max tokens & True \\
& Dtype & bfloat16 \\
\hline
\multirow{3}{*}{Agent}
& Max iterations per feature& 10 \\
& Confidence threshold& 0.75 \\
& Min queries per feature & 5 \\
\hline
\multirow{5}{*}{Retrieval}
& Embedding model& paraphrase-multilingual-MiniLM-L12-v2 \\
& Embedding batch size & 64 \\
& BM25 $k_1$ & 1.5 \\
& BM25 $b$ & 0.75 \\
& Hybrid $\alpha$ (dense weight) & 0.5 \\
& MMR $\lambda$& 0.6 \\
\hline
\end{tabular}
\caption{System hyperparameters. The agent uses a ReAct loop capped at 10 iterations per feature, with a confidence threshold of 0.75 for accepting conclusions. Retrieval combines BM25 and dense embeddings with equal weight ($\alpha=0.5$), followed by MMR reranking ($\lambda=0.6$) for diversity.}
\label{tab:hyperparameters}
\end{table*}

\section{Grambank Domain Types and Distributions}
\label{domain-grouping}
Table \ref{tab:grambank-features-domain} shows the detailed information about Grambank feature domain distribution according to WALS.
\begin{table*}[t]
\centering
\resizebox{\textwidth}{!}{
\begin{tabular}{lp{13cm}}
\toprule
\textbf{Domain} & \textbf{Feature IDs} \\
\midrule
Word Order (10) & \texttt{GB074}, \texttt{GB075}, \texttt{GB130}, \texttt{GB131}, \texttt{GB132}, \texttt{GB133}, \texttt{GB134}, \texttt{GB135}, \texttt{GB136}, \texttt{GB260} \\
\addlinespace[2pt]
Simple Clauses (66) & \texttt{GB068}, \texttt{GB070}, \texttt{GB071}, \texttt{GB072}, \texttt{GB073}, \texttt{GB089}, \texttt{GB090}, \texttt{GB091}, \texttt{GB092}, \texttt{GB093}, \texttt{GB094}, \texttt{GB095}, \texttt{GB096}, \texttt{GB098}, \texttt{GB099}, \texttt{GB103}, \texttt{GB104}, \texttt{GB105}, \texttt{GB107}, \texttt{GB108}, \texttt{GB109}, \texttt{GB113}, \texttt{GB114}, \texttt{GB115}, \texttt{GB116}, \texttt{GB117}, \texttt{GB126}, \texttt{GB137}, \texttt{GB138}, \texttt{GB139}, \texttt{GB140}, \texttt{GB147}, \texttt{GB148}, \texttt{GB149}, \texttt{GB155}, \texttt{GB156}, \texttt{GB250}, \texttt{GB252}, \texttt{GB253}, \texttt{GB254}, \texttt{GB256}, \texttt{GB257}, \texttt{GB262}, \texttt{GB263}, \texttt{GB264}, \texttt{GB265}, \texttt{GB266}, \texttt{GB270}, \texttt{GB273}, \texttt{GB275}, \texttt{GB276}, \texttt{GB285}, \texttt{GB286}, \texttt{GB291}, \texttt{GB297}, \texttt{GB302}, \texttt{GB303}, \texttt{GB304}, \texttt{GB322}, \texttt{GB323}, \texttt{GB324}, \texttt{GB401}, \texttt{GB408}, \texttt{GB409}, \texttt{GB410}, \texttt{GB522} \\
\addlinespace[2pt]
Nominal Categories (44) & \texttt{GB020}, \texttt{GB021}, \texttt{GB028}, \texttt{GB030}, \texttt{GB031}, \texttt{GB035}, \texttt{GB036}, \texttt{GB037}, \texttt{GB038}, \texttt{GB039}, \texttt{GB041}, \texttt{GB042}, \texttt{GB043}, \texttt{GB044}, \texttt{GB046}, \texttt{GB051}, \texttt{GB052}, \texttt{GB053}, \texttt{GB054}, \texttt{GB057}, \texttt{GB058}, \texttt{GB165}, \texttt{GB166}, \texttt{GB167}, \texttt{GB184}, \texttt{GB185}, \texttt{GB186}, \texttt{GB187}, \texttt{GB188}, \texttt{GB192}, \texttt{GB196}, \texttt{GB197}, \texttt{GB198}, \texttt{GB313}, \texttt{GB314}, \texttt{GB315}, \texttt{GB316}, \texttt{GB317}, \texttt{GB318}, \texttt{GB319}, \texttt{GB320}, \texttt{GB321}, \texttt{GB325}, \texttt{GB415} \\
\addlinespace[2pt]
Nominal Syntax (23) & \texttt{GB022}, \texttt{GB023}, \texttt{GB024}, \texttt{GB025}, \texttt{GB026}, \texttt{GB027}, \texttt{GB059}, \texttt{GB065}, \texttt{GB069}, \texttt{GB170}, \texttt{GB171}, \texttt{GB172}, \texttt{GB193}, \texttt{GB203}, \texttt{GB204}, \texttt{GB301}, \texttt{GB305}, \texttt{GB306}, \texttt{GB326}, \texttt{GB430}, \texttt{GB431}, \texttt{GB432}, \texttt{GB433} \\
\addlinespace[2pt]
Verbal Categories (19) & \texttt{GB082}, \texttt{GB083}, \texttt{GB084}, \texttt{GB086}, \texttt{GB110}, \texttt{GB111}, \texttt{GB119}, \texttt{GB120}, \texttt{GB121}, \texttt{GB146}, \texttt{GB177}, \texttt{GB298}, \texttt{GB299}, \texttt{GB309}, \texttt{GB312}, \texttt{GB400}, \texttt{GB519}, \texttt{GB520}, \texttt{GB521} \\
\addlinespace[2pt]
Morphology (11) & \texttt{GB047}, \texttt{GB048}, \texttt{GB049}, \texttt{GB079}, \texttt{GB080}, \texttt{GB081}, \texttt{GB122}, \texttt{GB124}, \texttt{GB158}, \texttt{GB159}, \texttt{GB160} \\
\addlinespace[2pt]
Complex Sentences (11) & \texttt{GB118}, \texttt{GB150}, \texttt{GB151}, \texttt{GB152}, \texttt{GB327}, \texttt{GB328}, \texttt{GB329}, \texttt{GB330}, \texttt{GB331}, \texttt{GB421}, \texttt{GB422} \\
\addlinespace[2pt]
Lexicon (11) & \texttt{GB123}, \texttt{GB127}, \texttt{GB129}, \texttt{GB296}, \texttt{GB300}, \texttt{GB333}, \texttt{GB334}, \texttt{GB335}, \texttt{GB336}, \texttt{GB402}, \texttt{GB403} \\
\addlinespace[2pt]
\bottomrule
\end{tabular}
}
\caption{Grambank features used in this study, grouped by domain (195 total).}
\label{tab:grambank-features-domain}
\end{table*}

\section{Sample \tfc{}Agent Traces}
\label{grambank-agent-trace-example}
Figure~\ref{fig:agent-trace-gb049-error} shows a complete agent trace for coding Grambank feature GB049 in Japhug, including the retrieved grammar sections, intermediate evidence, final label prediction, and audit result.

\begin{figure*}[t]
\centering

\begin{tcolorbox}[steptcb,
colback=lightblue, colframe=queryblue]
\textcolor{queryblue}{\textbf{Query}} \hfill
{\ttfamily\footnotesize GB049 $\cdot$ NMZObject $\cdot$ Coding$\{0,1,?\}$}\\[1pt]
\textit{Is there a productive morphological pattern for deriving an object noun from a verb?}\\[2pt]
{\footnotesize\textcolor{gray}{Phenomena:
\textrm{[object nominalization, noun--verb compounds, derivational morphology]}}}
\end{tcolorbox}

\vspace{-14pt}

\begin{tcolorbox}[steptcb,
colback=lightgreen, colframe=toolgreen]
\textcolor{toolgreen}{\textbf{Iter\,1}} \quad
{\ttfamily search\_text("object nominal patient derivation", top\_k=5)}\\[2pt]
\textbf{Thought:} The current evidence is contradictory regarding the productivity
of nominalization prefixes and their semantic scope (action vs. object).
Claim C003 suggests they form action nominals, while the query asks for object
nouns. Claim C001 suggests Object--Verb compounds are productive, but it is
unclear if these are considered ``derived from a verb'' in the strict
morphological sense or just compounding.\\[1pt]
\textcolor{gray}{\textit{Finding:}
This will identify if there is a specific derivational process for object nouns
distinct from action nominals or compounds.}
\end{tcolorbox}

\vspace{-14pt}

\begin{tcolorbox}[steptcb,
colback=lightgreen, colframe=toolgreen]
\textcolor{toolgreen}{\textbf{Iter\,2}} \quad
{\ttfamily extract\_author\_claims("Nominal morphology > Nominal derivations")}\\[2pt]
\textbf{Thought:} The current evidence is contradictory and low confidence.
Supporting claims suggest Object--Verb compounds are productive, while
contradicting claims suggest nominalizing prefixes are not fully productive.
Crucially, the query asks for deriving an \textit{object} noun from a verb.\\[1pt]
\textcolor{gray}{\textit{Finding:}
This will reveal if the grammar describes a specific derivational process for
creating object/patient nouns from verbs, distinct from compounding or action
nominalization.}
\end{tcolorbox}

\vspace{-14pt}

\begin{tcolorbox}[steptcb,
colback=gray!8, colframe=gray!60]
\centering
\textcolor{gray!70}{\textbf{\large $\vdots$}}\\[-1pt]
{\footnotesize\textcolor{gray}{
Repeated calls }}
\end{tcolorbox}

\vspace{-14pt}

\begin{tcolorbox}[steptcb,
colback=lightgreen, colframe=toolgreen]
\textcolor{toolgreen}{\textbf{Iter\,10}} \quad
{\ttfamily extract\_author\_claims("Nominal morphology > Nominal derivations")}\\[2pt]
\textbf{Thought:} The evidence so far indicates that Object--Verb compounds
are productive (C001), but the query asks about deriving an object noun
\textit{from a verb}. Compounding involves a noun and a verb, not deriving a
noun solely from a verb. The prefixation of nominalizing prefixes creates
action nominals or abstract nouns, not necessarily object nouns
(patient/result).\\[1pt]
\textcolor{gray}{\textit{Finding:}
This will reveal if the grammar describes a specific derivational process for
creating object nouns from verbs, distinct from compounding or action
nominalization.}
\end{tcolorbox}

\vspace{-14pt}

\begin{tcolorbox}[steptcb,
colback=lightred, colframe=concludered]
\textcolor{concludered}{\textbf{Conclude}} \quad
{\ttfamily predict\_label}\\[2pt]
\textbf{Label reasoning:} The evidence indicates that object nominalization is
primarily achieved through compounding (Object--Verb compounds) rather than a
specific morphological element attaching to verbs to form an object
nominalization; the available affixes form action/abstract nominals and are
not fully productive for object derivation.\\[2pt]
\textbf{Answer:} In Japhug, there is no single, fully productive morphological
affix dedicated exclusively to deriving object nouns from verbs. Instead, the
language primarily utilizes a highly productive syntactic pattern of
Object--Verb nominal compounds to express this relationship.
\end{tcolorbox}

\vspace{-14pt}

\begin{tcolorbox}[steptcb,
colback=lightyellow, colframe=outputgold]
\textcolor{outputgold}{\textbf{Output}} \hfill
{\footnotesize\textcolor{gray}{27 LLM calls $\cdot$ 46{,}228 tokens
 (40{,}366 in / 5{,}862 out)}}\\[4pt]
\begin{tabular}{@{}l@{\quad}l@{\qquad}l@{\quad}l@{}}
\textbf{Label:} & \texttt{0} &
\textbf{Confidence:} & \texttt{0.10} \\[3pt]
\textbf{Audit:} & \texttt{overturned} &
\textbf{Human review:} & \texttt{true} \\
\end{tabular}\\[3pt]
\textbf{Review reason:}
The answer is factually incorrect as the evidence explicitly identifies
productive Object--Verb nominal compounds.\\[2pt]
\textbf{Audit objections:}
Evidence cites productive OV compounds, contradicting ``no pattern'';
answer ignores supporting claims C001--C003.
\end{tcolorbox}

\caption{Agent trace for Grambank feature \textbf{GB049} (\textit{NMZObject}) in Japhug, produced by the grammar-grounded
pipeline with Qwen3.6-27B. 
The trace shows that the system predicts label \texttt{0} with low confidence, flags the case
for human review, and the audit overturns the conclusion.}
\label{fig:agent-trace-gb049-error}
\end{figure*}

\section{Sample \tht{}Agent Traces}
Figure \ref{fig:agent-trace-u30} shows the agent trace results for testing Greenberg Universal U30 with Qwen3.6-27B under Grammar setting.
\label{hypo-agent-trace}

\begin{figure*}[t]
\centering

\begin{tcolorbox}[steptcb,
colback=lightblue, colframe=queryblue]
\textcolor{queryblue}{\textbf{Query}} \hfill
{\ttfamily\footnotesize U30 $\cdot$ Implication}\\[1pt]
\textit{If the verb has categories of person-number or of gender, it always has tense-mode categories.}\\[2pt]
{\footnotesize\textcolor{gray}{Domain:
\texttt{TENSE\_ASPECT\_MODALITY};
logic: antecedent $\Rightarrow$ consequent;
languages: \texttt{10}}}
\end{tcolorbox}

\vspace{-14pt}

\begin{tcolorbox}[steptcb,
colback=lightgreen, colframe=toolgreen]
\textcolor{toolgreen}{\textbf{Plan}} \quad
{\ttfamily parse\_universal(U30)}\\[2pt]
\textbf{Antecedent:} verb has categories of person-number or of gender.\\[1pt]
\textbf{Consequent:} verb also has tense-mode categories.\\[1pt]
\textcolor{gray}{\textit{Parsing rationale:} from CSV}
\end{tcolorbox}

\vspace{-14pt}

\begin{tcolorbox}[steptcb,
colback=lightgreen, colframe=toolgreen]
\textcolor{toolgreen}{\textbf{Language assessments}} \quad
{\ttfamily aggregate\_language\_verdicts}\\[2pt]

\begin{center}
\begin{tabular}{ll}
\hline
\textbf{Assessment} & \textbf{Languages} \\
\hline
\texttt{VALID\_SUPPORT} &
Aguaruna, Ch\'{a}cobo, Moloko, Komnzo, Pite Saami \\
\texttt{VALID\_VIOLATION} &
Ik \\
\texttt{ANTECEDENT\_NA} &
Papuan Malay, Kalamang \\
\texttt{INSUFFICIENT} &
Chakali, Eastern Geshiza \\
\hline
\end{tabular}
\end{center}

\textcolor{gray}{\textit{Counts:}
support = \texttt{5};
violation = \texttt{1};
antecedent N/A = \texttt{2};
insufficient = \texttt{2}.}
\end{tcolorbox}

\vspace{-14pt}

\begin{tcolorbox}[steptcb,
colback=lightgreen, colframe=toolgreen]
\textcolor{toolgreen}{\textbf{Counterexample}} \quad
{\ttfamily inspect\_language("Ik")}\\[2pt]
\textbf{Antecedent evidence:} Ik verbs are richly inflected for person and
number, utilizing a complex system of prefixes and suffixes for subject and
object agreement.\\[1pt]
\textbf{Consequent evidence:} Ik lacks a dedicated, obligatory tense paradigm;
temporal distinctions are conveyed through aspectual markers, auxiliaries,
or context rather than fused tense-mode.\\[1pt]
\textcolor{gray}{\textit{Assessment:}
\texttt{VALID\_VIOLATION};
confidence = \texttt{0.85}.}
\end{tcolorbox}

\vspace{-14pt}

\begin{tcolorbox}[steptcb,
colback=lightred, colframe=concludered]
\textcolor{concludered}{\textbf{Conclude}} \quad
{\ttfamily conclude}\\[2pt]
\textbf{Conclusion:} The universal is false because there is at least one valid
violation where a language exhibits person-number agreement on the verb but
lacks obligatory tense-mode marking. While many languages support the
correlation, the existence of counterexamples like Ik demonstrates that
person-number agreement does not strictly necessitate tense-mode categories.\\[2pt]
\textbf{Cross-linguistic notes:} Ik serves as a critical counterexample.
In contrast, Aguaruna, Ch\'{a}cobo, Moloko, Komnzo, and Pite Saami support the
universal by exhibiting both features. Papuan Malay and Kalamang are irrelevant
to the test because they lack the requisite person-number agreement on verbs.
\end{tcolorbox}

\vspace{-14pt}

\begin{tcolorbox}[steptcb,
colback=lightyellow, colframe=outputgold]
\textcolor{outputgold}{\textbf{Output}} \hfill
{\footnotesize\textcolor{gray}{10 languages evaluated}}\\[4pt]
\begin{tabular}{@{}l@{\quad}l@{\qquad}l@{\quad}l@{}}
\textbf{Verdict:} & \texttt{FALSE} &
\textbf{Confidence:} & \texttt{0.95} \\[3pt]
\textbf{Support:} & \texttt{5} &
\textbf{Violation:} & \texttt{1} \\[3pt]
\textbf{Antecedent N/A:} & \texttt{2} &
\textbf{Insufficient:} & \texttt{2} \\
\end{tabular}
\end{tcolorbox}

\caption{Agent trace for typological hypothesis testing on Greenberg Universal
\textbf{U30}. The system parses the universal as an implication, evaluates the
antecedent and consequent across 10 languages, identifies Ik as a valid
counterexample, and concludes that the universal is false with confidence
\texttt{0.95}.}
\label{fig:agent-trace-u30}
\end{figure*}

\section{Agent Prompts}
\label{app:alltemp}
Figures~\ref{fig:prompts-ab}, \ref{fig:prompts-cd}, \ref{fig:prompts-ef}, and \ref{fig:prompts-gh} illustrate representative prompt templates used at different stages of the system workflow.

\lstdefinestyle{prompt}{
basicstyle=\ttfamily\scriptsize,
breaklines=true,
breakatwhitespace=false,
frame=single,
framesep=3pt,
xleftmargin=3pt,
xrightmargin=3pt,
columns=fullflexible,
keepspaces=true,
belowskip=0pt,
aboveskip=0pt,
}

\begin{figure*}[t]
\begin{minipage}[t]{0.49\textwidth}
\textbf{(a) Domain Extraction Prompt}
\vspace{2pt}
\begin{lstlisting}[style=prompt]
You are a linguistic typologist. Below is the
table of contents of a reference grammar for
{language} with section summaries, and
quantitative IGT corpus statistics.
 
TABLE OF CONTENTS WITH SECTION SUMMARIES:
{toc_with_summaries}
 
QUANTITATIVE IGT SUMMARY:
{igt_summary}
 
GLOSS ABBREVIATION LEGEND:
{abbrev_legend}
 
Task: Identify major typological domains and
generate 3-5 specific, answerable feature
questions per domain grounded in (a) section
summaries and (b) IGT tag frequencies.
 
Heuristics:
- PST/PFV frequent -> generate TMA questions
- EVID at 0% -> EVIDENTIALITY likely absent
- No agreement tags -> AGREEMENT may be "No"
 
Output ONLY valid JSON:
{
"domains": [{
"domain_id": "D001",
"domain_name": "TENSE_ASPECT_MODALITY",
"relevant_sections": ["Verbs > Tense"],
"igt_signals": ["PST: 140 (8.4%)"],
"candidate_features": [{
"feature_id": "F001",
"question": "Does {language} grammaticalize
 tense?",
"igt_tags_to_check": ["PST","FUT","PRF"],
"prior": "likely_yes"
}]
}]
}
\end{lstlisting}
\end{minipage}
\hfill
\begin{minipage}[t]{0.49\textwidth}
\textbf{(b) ReAct Action-Selection Prompt}
\vspace{2pt}
\begin{lstlisting}[style=prompt]
You are deep-searching a reference grammar
for a typological feature.
 
LANGUAGE: {language}
FEATURE QUESTION: {question}
CURRENT EVIDENCE GRAPH: {evidence_summary}
EVIDENCE GAPS: {gap_analysis}
Progress: iteration {iteration}/{max_iter}
 
AVAILABLE TOOLS:
 1. read_full_section(query)
 2. follow_cross_references(query)
 3. extract_author_claims(query)
 4. search_text(query, top_k)
 5. analyse_tag(tag)
 6. analyse_construction(tags)
 7. analyse_absence(category)
 8. compare_tags(tag_a, tag_b)
 9. get_section_igt(query)
10. search_translations(query)
11. get_triline_examples(query)
12. conclude[only when ALL constraints met]
 
Gap-to-tool guidance:
- NO IGT EVIDENCE-> analyse_tag / absence
- NO GRAMMAR PROSE -> read_full_section
- CONTRADICTIONS -> follow_cross_references
 
Output ONLY valid JSON:
{
"thought": "what gap this action fills",
"action": "tool_name",
"args": {},
"evidence_type": "grammar_prose|igt_quant|...",
"claim_to_add": "one-sentence claim",
"supports_hypothesis": true | false | null
}
\end{lstlisting}
\end{minipage}
\caption{(a) Domain extraction prompt: the model reads grammar section summaries and IGT
tag frequencies to discover typological domains and generate candidate feature questions.
(b) ReAct action-selection prompt: at each iteration the agent receives its evidence graph
and gap analysis, then selects one of 11 tools; \texttt{conclude} is permitted only when
all required evidence types are present.}
\label{fig:prompts-ab}
\end{figure*}

\begin{figure*}[t]
\begin{minipage}[t]{0.49\textwidth}
\textbf{(c) Contradiction Resolution Prompt}
\vspace{2pt}
\begin{lstlisting}[style=prompt]
You are resolving a contradiction in the
evidence for a typological feature.
 
LANGUAGE: {language}
FEATURE QUESTION: {question}
 
CLAIM A: {claim_a_text}
Source: {claim_a_source}
 
CLAIM B: {claim_b_text}
Source: {claim_b_source}
 
Consider:
- Is one source more authoritative?
- Could both be true (variation, different
construction types)?
- Does one apply to a restricted context
the other does not mention?
 
Output ONLY valid JSON:
{
"resolution": "how contradiction is resolved",
"resolved": true | false,
"preferred_claim": "A"|"B"|"both_partial"
 |"neither",
"revised_confidence_a": 0.0-1.0,
"revised_confidence_b": 0.0-1.0,
"synthesis": "one sentence: nuanced truth"
}
\end{lstlisting}
\end{minipage}
\hfill
\begin{minipage}[t]{0.49\textwidth}
\textbf{(d) Post-hoc Auditor Prompt}
\vspace{2pt}
\begin{lstlisting}[style=prompt]
You are a critical auditor reviewing a
typological conclusion.
 
LANGUAGE: {language}
FEATURE: {question}
CONCLUSION: {value} -- "{value_detail}"
CONFIDENCE: {confidence}
EVIDENCE GRAPH SUMMARY:
{evidence_graph_summary}
 
Try to DISPROVE this conclusion using only
the evidence listed:
- Are IGT patterns diagnostic or correlated?
- Could grammar prose describe marginal usage?
- Are contradictions ignored in conclusion?
- Should "Yes" be "Partial" given evidence?
 
Output ONLY valid JSON:
{
"verdict": "upheld"|"weakened"|"overturned",
"objections": [
"objection 1 (max 60 chars)",
"objection 2 (max 60 chars)"
],
"revised_value": "same or corrected",
"revised_confidence": 0.0-1.0,
"audit_notes": "one sentence summary"
}
\end{lstlisting}
\end{minipage}
\caption{(c) Contradiction resolution prompt: invoked when the evidence graph detects a
conflict between two claims; the ReAct loop pauses until resolution is complete.
(d) Post-hoc auditor prompt: an adversarial pass that attempts to overturn or weaken the
conclusion; the verdict and revised confidence are stored in the final feature record.}
\label{fig:prompts-cd}
\end{figure*}

\begin{figure*}[t]
\begin{minipage}[t]{0.49\textwidth}
\textbf{(e) Feature Conclusion Prompt}
\vspace{2pt}
\begin{lstlisting}[style=prompt]
You are writing a final typological feature
entry.
 
LANGUAGE: {language}
FEATURE QUESTION: {question}
COMPLETE EVIDENCE GRAPH: {evidence_graph}
CHUNK SUMMARIES: {chunk_summaries}
 
Evidence weighting:
- GRAMMAR_STATEMENT > IGT_PATTERN alone
- ABSENCE_EVIDENCE with zero counts =
strong negative evidence
- Unresolved contradictions -> lower
confidence, flag for review
- "Yes" requires BOTH grammar prose AND
IGT quantitative support
 
Output ONLY valid JSON:
{
"linguistic_definition": "lang-agnostic
definition (1-2 sentences)",
"structural_description": "realization
in {language}: forms, positions",
"value": "Yes|No|Partial|Unclear|?",
"value_detail": "key markers + refs",
"confidence": 0.0-1.0,
"key_evidence": [
"Finding: ... Source: Sec [chunk_id].
 Justification: ...",
"Finding: PST 140/1668 (8.4%), preverbal.
 Source: analyse_tag(PST)."
],
"typological_notes": "cross-ling. notes",
"needs_human_review": true | false,
"review_reason": "one sentence or empty"
}
\end{lstlisting}
\end{minipage}
\hfill
\begin{minipage}[t]{0.49\textwidth}
\textbf{(f) Grambank Labeler Prompt}
\vspace{2pt}
\begin{lstlisting}[style=prompt]
You are a typology expert applying a
Grambank coding scheme to a grammar.
 
FEATURE: {grambank_id} -- {query}
CODING POLICY (apply EXACTLY):
{coding_policy}
 
EVIDENCE (language: {language})
Answer summary: {answer}
Key evidence items: {key_evidence}
Structural description: {structural_desc}
IGT examples cited: {igt_examples}
 
Task:
1. Note what grammatical behaviour each
 piece of evidence demonstrates.
2. Match evidence to coding criteria.
3. Select the label most directly
 supported by a concrete IGT example
 or explicit grammar statement.
 
Rules:
- Label must be exactly one code from
the Coding Policy (e.g. 0, 1, 2, ?).
- Use "?" only when evidence cannot
distinguish between substantive codes.
 
Output ONLY valid JSON:
{
"label": "<code from Coding Policy>",
"reasoning": "2-4 sentences citing
specific evidence items",
"confidence": 0.0-1.0
}
\end{lstlisting}
\end{minipage}
\caption{(e) Feature conclusion prompt: synthesizes the full evidence graph into a
structured feature record; evidence items must cite grammar sections by chunk ID and
IGT tools by name.
(f) Grambank labeler prompt: maps a \texttt{QueryResult} to a Grambank categorical code
by matching retrieved evidence against the official coding policy.}
\label{fig:prompts-ef}
\end{figure*}

\begin{figure*}[t]
\begin{minipage}[t]{0.49\textwidth}
\textbf{(g) IGT-Only Domain Extraction Prompt}
\vspace{2pt}
\begin{lstlisting}[style=prompt]
You are a linguistic typologist. There is
NO reference grammar. Infer typological
domains purely from IGT corpus statistics.
 
LANGUAGE: {language}
IGT CORPUS STATISTICS: {igt_summary}
CONSTRUCTION PATTERNS: {construction_patterns}
GLOSS ABBREVIATION LEGEND: {abbrev_legend}
 
Reading the data:
- Tags >5% coverage-> likely grammaticalized
- Tags <0.5% -> marginal or absent
- Consistent position -> morphosyntactic slot
- Complementary tags (PST/FUT, SG/PL)
-> paradigm systems
 
Procedure:
1. Group tags by domain (TMA, Agreement,
 Case, Negation, ...)
2. For each domain: PRESENT (>5%),
 PARTIAL (<5%), ABSENT (near-zero)
3. Generate 3-5 feature questions per domain
4. Set prior: "likely_yes" / "likely_no"
/ "uncertain"
 
Output ONLY valid JSON:
{
"domains": [{
"domain_id": "D001",
"domain_name": "TENSE_ASPECT_MODALITY",
"igt_signals": ["PST: 140 (8.4%)",
"FUT: 0 (0.0%)"],
"candidate_features": [{
"feature_id": "F001",
"question": "Does {language}
 grammaticalize tense?",
"igt_tags_to_check": ["PST","FUT"],
"prior": "likely_yes"
}]
}]
}
\end{lstlisting}
\end{minipage}
\hfill
\begin{minipage}[t]{0.49\textwidth}
\textbf{(h) IGT-Only ReAct Decision Prompt}
\vspace{2pt}
\begin{lstlisting}[style=prompt]
You are deep-searching an IGT corpus for a
typological feature. There is NO grammar.
 
LANGUAGE: {language}
FEATURE QUESTION: {question}
CURRENT EVIDENCE GRAPH: {evidence_summary}
EVIDENCE GAPS: {gap_analysis}
Progress: iteration {iteration}/{max_iter}
 
AVAILABLE TOOLS (IGT-only):
 1. get_tag_inventory
 2. get_construction_inventory
 3. find_tag_cluster(seed_tag)
 4. analyse_tag(tag)
 5. analyse_construction(tags)
 6. analyse_absence(category)
 7. compare_tags(tag_a, tag_b)
 8. get_section_igt(query)
 9. analyse_semantic_context(tag)
10. search_translations(query)
11. get_triline_examples(query)
12. analyse_morpheme_position(tag)
13. parse_example_structure(query)
14. get_morpheme_forms(tag)
15. conclude [only when ALL constraints met]
 
Gap-to-tool guidance:
- NO IGT EVIDENCE -> analyse_tag
- NO ABSENCE CHECK-> analyse_absence
- NO COUNTER-EV.-> compare_tags
- CONTRADICTIONS-> find_tag_cluster
 
Output ONLY valid JSON:
{
"thought": "what gap this action fills",
"action": "tool_name",
"args": {},
"evidence_type": "igt_quantitative|...",
"claim_to_add": "one-sentence claim",
"supports_hypothesis": true | false | null
}
\end{lstlisting}
\end{minipage}
\caption{(g) IGT-only domain extraction prompt: when no grammar is available, domain
discovery is driven entirely by corpus statistics; high-frequency tags signal
grammaticalized categories, complementary tags signal paradigms, and near-zero counts
signal absence.
(h) IGT-only ReAct decision prompt: mirrors the grammar-grounded variant but expands
the tool set to 15 IGT-specific tools including morpheme position analysis and
LLM-powered clause parsing.}
\label{fig:prompts-gh}
\end{figure*}

\section{Additional Statistical Analysis}
\label{app:statistical_analysis}

We conduct a language-clustered bootstrap analysis with 10,000 resamples.
Languages were treated as the resampling units to account for within-language
correlation across typological features. We report paired comparisons between information
conditions. P-values are Holm-corrected within each model and metric across
the three pairwise comparisons. Results are shown in Table~\ref{tab:pairwise_ci}.

\begin{table*}
\centering
\begin{adjustbox}{width=\textwidth}
\begin{tabular}{llccc|ccc}
\hline
\textbf{Model} & \textbf{A vs.\ B} &
\multicolumn{3}{c|}{\textbf{Weighted F1}} &
\multicolumn{3}{c}{\textbf{Macro F1}} \\
& &
$\Delta$ & 95\% CI & $p$ (Holm) &
$\Delta$ & 95\% CI & $p$ (Holm) \\
\hline
Gemma4-31B
& Grammar $-$ IGT
& 4.59 & [2.58, 6.57] & $<.001$
& 7.60 & [1.49, 12.44] & .024 \\

& Grammar+IGT $-$ Grammar
& $-2.81$ & [$-5.62$, 0.18] & .128
& $-1.52$ & [$-6.09$, 2.41] & .433 \\

& Grammar+IGT $-$ IGT
& 1.78 & [$-1.04$, 4.54] & .224
& 6.08 & [0.09, 10.45] & .094 \\
\hline

GPT-5.4-mini
& Grammar $-$ IGT
& 9.58 & [5.62, 13.54] & $<.001$
& 12.60 & [6.74, 17.53] & $<.001$ \\

& Grammar+IGT $-$ Grammar
& $-2.09$ & [$-4.37$, 0.60] & .118
& $-0.64$ & [$-5.66$, 4.80] & .831 \\

& Grammar+IGT $-$ IGT
& 7.49 & [3.97, 11.07] & $<.001$
& 11.97 & [8.22, 15.25] & $<.001$ \\
\hline

Qwen2.5-7B
& Grammar $-$ IGT
& 3.11 & [1.85, 4.34] & $<.001$
& 12.89 & [5.17, 18.04] & $<.001$ \\

& Grammar+IGT $-$ Grammar
& $-5.44$ & [$-6.94$, $-3.83$] & $<.001$
& $-5.51$ & [$-10.83$, 0.95] & .097 \\

& Grammar+IGT $-$ IGT
& $-2.32$ & [$-4.00$, $-0.55]$ & .011
& 7.38 & [2.02, 12.45] & .011 \\
\hline

Qwen3.5-9B
& Grammar $-$ IGT
& 5.43 & [3.38, 7.22] & $<.001$
& 13.08 & [5.67, 19.50] & .005 \\

& Grammar+IGT $-$ Grammar
& $-7.73$ & [$-9.55$, $-5.80$] & $<.001$
& $-7.02$ & [$-12.76$, $-0.41$] & .074 \\

& Grammar+IGT $-$ IGT
& $-2.30$ & [$-4.69$, $-0.11$] & .040
& 6.05 & [$-0.94$, 12.40] & .088 \\
\hline

Qwen3.6-27B
& Grammar $-$ IGT
& $-0.83$ & [$-2.82$, 1.37] & 1
& 6.54 & [$-2.02$, 14.91] & .372 \\

& Grammar+IGT $-$ Grammar
& 0.79 & [$-1.39$, 2.82] & 1
& $-2.31$ & [$-7.82$, 3.19] & .421 \\

& Grammar+IGT $-$ IGT
& $-0.04$ & [$-2.56$, 2.38] & 1
& 4.23 & [$-1.04$, 9.95] & .372 \\
\hline
\end{tabular}
\end{adjustbox}
\caption{Pairwise differences between information conditions based on
language-clustered bootstrap resampling. $\Delta$ denotes the difference
in F1 between the two conditions. P-values are Holm-corrected within
each model and metric.}
\label{tab:pairwise_ci}
\end{table*}

\section{Iteration Influence on IGT-only Setting}
\label{app:iteration}
We compare reason iterations from 10 to 40 on the IGT-only setting of \texttt{Qwen2.5-7B}. Due to time limit, we implement experiments on 6 languages including Aguaruna, Bargam, Chakali, Ch\'{a}cobo, Hup, and Ik. Results are in Table \ref{tab:iteration_results}. While weighted F1 remains relatively stable, Macro F1 consistently improves with a larger reasoning budget, reaching its highest value at 40 iterations.

\begin{table*}
\centering
\begin{tabular}{c|cc}
\hline
\textbf{Setting} & \textbf{Weighted F1} & \textbf{Macro F1} \\
\hline
IGT-10-iter & 50.9 & 27.2  \\
IGT-20-iter & 51.2 & 33.2 \\
IGT-30-iter & 49.3 & 35.1 \\
IGT-40-iter & 51.8 & 37.8 \\
\hline
\end{tabular}
\caption{Performance with different iteration settings.}
\label{tab:iteration_results}
\end{table*}

\section{Ablation of the Auditing Stage}
\label{app:audit}
Table~\ref{tab:audit_ablation} shows the estimated contribution of the auditing stage on Aguaruna. We conduct this ablation under the Grammar-only setting. Overall, auditing provides consistent gains across the three models, although the magnitude of improvement varies by model.
\begin{table*}[t]
\centering
\resizebox{\textwidth}{!}{
\begin{tabular}{l|ccc|ccc}
\hline
\textbf{Model} &
\textbf{W-F1 w/o Audit} &
\textbf{W-F1 w/ Audit} &
\textbf{$\Delta$ W-F1} &
\textbf{M-F1 w/o Audit} &
\textbf{M-F1 w/ Audit} &
\textbf{$\Delta$ M-F1} \\
\hline
GPT5.4-mini  & 67.2 & 67.5 & +0.3 & 68.3 & 68.5 & +0.2 \\
Gemma4-31B & 67.0 & 69.5 & +2.5 & 68.5 & 69.9 & +1.4 \\
Qwen3.6-27B  & 58.6 & 61.1 & +2.5 & 69.6 & 70.9 & +1.3 \\
\hline
\end{tabular}
}
\caption{Estimated effect of the auditing stage on Aguaruna. Metrics includes weighted F1 and macro F1.}
\label{tab:audit_ablation}
\end{table*}

\end{document}